\documentclass[11pt, a4paper, logo, copyright, nonumbering]{xiaomi}

\usepackage[numbers, sort&compress, square]{natbib}
\usepackage{dblfloatfix}
\usepackage[normalem]{ulem}
\usepackage{caption}
\usepackage{dramatist}
\usepackage{xspace}
\usepackage{pifont} 
\usepackage{multirow}
\usepackage{tcolorbox}
\usepackage{xltabular}
\usepackage{longtable}
\usepackage{hyperref}
\usepackage{wrapfig}

\usepackage{amsfonts}
\usepackage{amsmath}
\usepackage{amssymb}
\usepackage{lineno}
\usepackage{multirow}
\usepackage{adjustbox}

\usepackage[bottom]{footmisc}

\usepackage{CJKutf8}
\usepackage{setspace}
\usepackage{makecell}
\usepackage{graphicx}
\usepackage{subcaption}
\usepackage{multicol} 

\usepackage[utf8]{inputenc} 
\usepackage[T1]{fontenc}    
\usepackage{hyperref}
\usepackage{cleveref}
\usepackage{url}            
\usepackage{booktabs}       
\usepackage{amsfonts}       
\usepackage{nicefrac}       
\usepackage{microtype}      
\usepackage{xcolor}         
\usepackage{colortbl}
\usepackage{tcolorbox}
\usepackage{xspace}
\definecolor{BrickRed}{rgb}{.72,0,0}
\definecolor{darkgreen}{rgb}{0.0, 0.5, 0.0}
\definecolor{ForestGreen}{RGB}{34,139,34}
\definecolor{LakeBlue}{RGB}{0,61,153}
\definecolor{MiOrange}{RGB}{255,225,204}
\definecolor{Hex}{RGB}{225,213,231}

\hypersetup{
    colorlinks=true,
    allcolors=LakeBlue
}

\title{\centering From Synthesis to Removal: Physics-Grounded Reflection Simulation and Diffusion-Based Video Dereflection}

\titlerunning{Physics-Grounded Reflection Simulation and Diffusion-Based Video Dereflection}

\author{
  Zepeng Wang\textsuperscript{*},
  Jiagao Hu\textsuperscript{*},
  Fuhao Li\textsuperscript{*},
  Yuxuan Chen,
  Fei Wang,
  Daiguo Zhou
}
\institute{
MiLM Plus, Xiaomi Inc.\\[3pt]
{\footnotesize\textsuperscript{*}Equal contribution.}
}

\begin{document}

\begin{abstract}
Videos captured through glass often contain reflections that degrade visual quality and interfere with downstream vision tasks.
Although single-image reflection removal has been extensively studied, video reflection removal remains largely underexplored due to the lack of paired video data, temporally coherent removal models, and dedicated evaluation benchmarks.
We present a closed-loop framework that unifies physics-grounded reflection simulation, diffusion-based video dereflection, and benchmark evaluation.
Our \textbf{S2R-Synthesis} pipeline generates paired reflected and reflection-free videos by performing physics-grounded augmentation in the structure space and rendering realistic reflected videos with a trained video diffusion renderer; the augmentation models key glass-related effects including roughness-induced blur, thickness-induced ghosting, and reflectance variation.
Based on the synthesized data, we introduce \textbf{S2R-Removal}, the first diffusion-based video reflection removal model, which adapts a pretrained video diffusion prior through reflection-aware latent adaptation and one-step pixel-geometric refinement, recovering the clean transmission in a single denoising step.
We further build \textbf{S2R-Bench}, the first benchmark for video reflection removal, supporting both full-reference evaluation and real-world human perceptual assessment.
Experiments on S2R-Bench and multiple public image benchmarks demonstrate state-of-the-art performance and faster inference than even non-diffusion baselines, and validate the effectiveness of S2R-Synthesis.
Project page: \url{https://codingwzp.github.io/VideoDereflection_S2R}.
\end{abstract}

\maketitle

\begin{figure*}
    \centering
    \includegraphics[width=\textwidth]{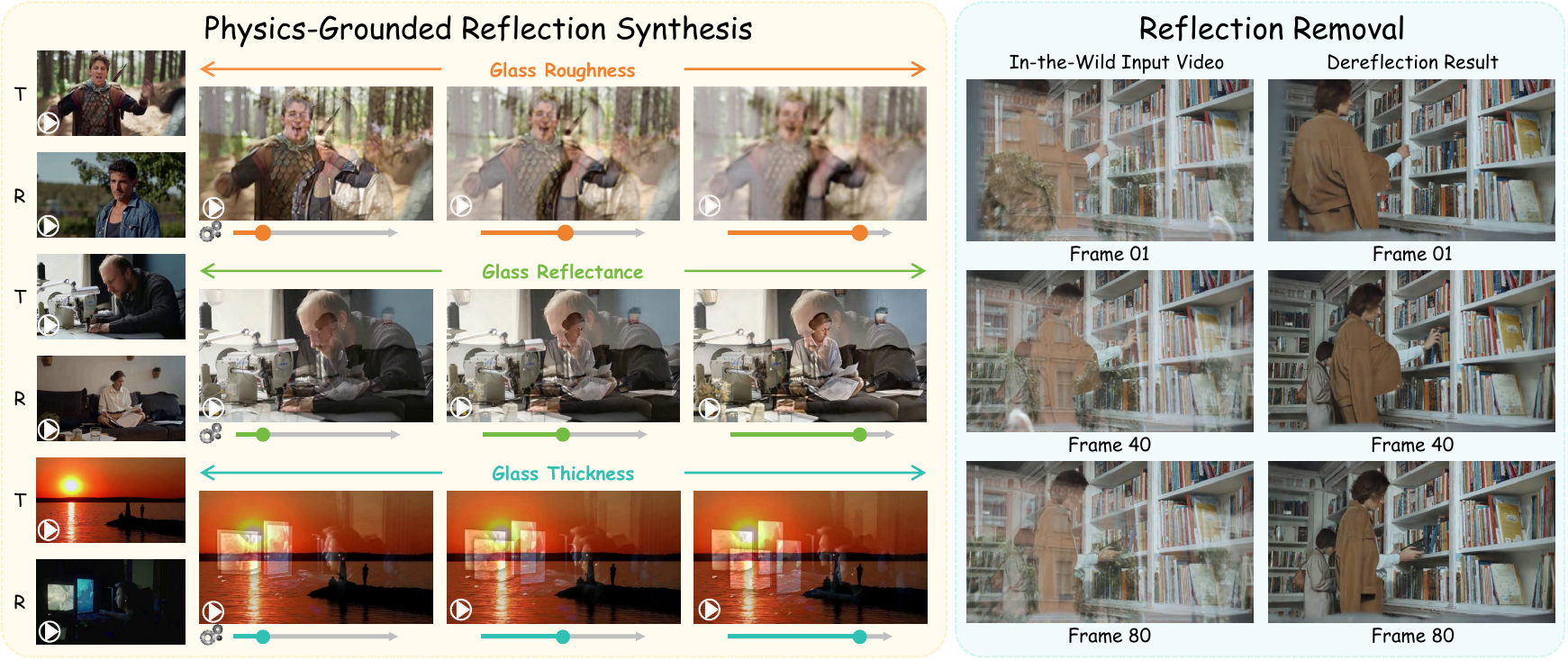}
    \caption{Overview of the reflection synthesis and removal pipeline.
    \textbf{Left:} Controllable synthesis across glass roughness, reflectance, and thickness.
    \textbf{Right:} Qualitative removal results on in-the-wild videos.}
    \label{fig:teaser}
\end{figure*}

\section{Introduction}
\label{sec:introduction}

Videos captured through glass are ubiquitous, yet the resulting reflections often obscure the underlying transmission scene, degrade perceptual quality, and interfere with downstream vision systems~\cite{farid1999separating,yang2016robust}.
Removing such reflections is therefore important for video restoration, computational photography, and practical visual perception.

Despite extensive progress in single-image reflection removal~\cite{zhang2018single,hu2023single,hu2026dereflection,li2025rectifying}, video reflection removal remains significantly underexplored.
Videos introduce additional challenges: reflections vary over time, move independently from the transmission layer, and interact with camera motion.
Applying image methods frame by frame is a straightforward solution but often produces flickering, inconsistent removal strength, and unstable background details due to the lack of temporal modeling~\cite{nandoriya2017video}.
A successful video dereflection method should instead suppress reflections while preserving the appearance, structure, and temporal coherence of the transmission video.

Achieving these goals, however, is bottlenecked by paired training data: obtaining a reflected video and its perfectly aligned reflection-free counterpart is extremely difficult, as camera motion, illumination, and dynamic scene content must remain consistent before and after reflection removal.
Existing synthesis strategies each fall short of providing scalable, realistic, and controllable reflections: RGB-space blending~\cite{hu2023single,wen2019single} yields aligned pairs but overly simplistic reflections that miss glass-dependent effects; frame-wise image diffusion~\cite{flux-2-2025} achieves realistic appearance per frame but introduces temporal flicker and offers no control over glass parameters; and direct video diffusion models, despite their temporal modeling capability, exhibit extremely low success rates when directly prompted to synthesize reflections on clean videos, often failing to produce visible reflections or corrupting the underlying scene content.

Beyond the data bottleneck, video reflection removal is further hampered by the absence of temporally coherent models and task-specific benchmarks.
We therefore approach it as a \emph{closed-loop synthesis-to-removal} problem---jointly tackling scalable paired video data, temporally coherent video dereflection models, and benchmarks for both controlled and real-world evaluation---rather than as an isolated model design problem.
To this end, we present a physics-grounded framework that unifies controllable reflection synthesis and reflection-aware video removal, complemented by a purpose-built benchmark for evaluation; representative results from both are illustrated in~\Cref{fig:teaser}.

First, we propose \textbf{S2R-Synthesis}, a paired video reflection synthesis pipeline that achieves both realism and controllable diversity.
S2R-Synthesis decouples reflection \emph{structure} from \emph{appearance}: a video diffusion renderer---distilled from a frozen image diffusion model for realism and inheriting temporal coherence from its video prior---renders photorealistic reflected videos from the clean transmission and a structured reflection condition, while Physics-Grounded Augmentation (PGA) transforms this condition according to glass optical parameters, providing control over major glass-related effects including surface roughness, glass reflectance, and thickness.

Second, we introduce \textbf{S2R-Removal}, a diffusion-based video reflection removal model trained on the synthesized paired data.
Since reflections attenuate rather than occlude the transmission layer, dereflection is fundamentally a restoration task rather than free generation.
Accordingly, S2R-Removal leverages the generative prior of video diffusion models~\cite{wan2025wan} through a two-stage training process: Stage~I performs reflection-aware latent adaptation with reflection-intensity supervision to localize and remove reflections, and Stage~II further refines the model with pixel-geometric losses (reconstruction, structural, and depth consistency), training it to recover the clean transmission in a single denoising step while preserving the underlying scene.
This one-step design further makes S2R-Removal substantially more efficient than even non-diffusion baselines.

Finally, we introduce \textbf{S2R-Bench}, the first benchmark for video reflection removal.
S2R-Bench contains two complementary subsets: S2R-Ref provides paired videos with clean ground truth for full-reference evaluation, while S2R-Real contains in-the-wild reflection videos for human perceptual assessment.
Together, they support evaluation of reconstruction fidelity, temporal consistency, reflection removal quality, and transmission preservation.

Our contributions are summarized as follows:
\begin{itemize}[leftmargin=1em, nosep]
    \item We present, to the best of our knowledge, \textit{the first closed-loop framework for video reflection removal}, integrating physics-grounded synthesis, diffusion-based removal, and benchmark evaluation.

    \item We propose \textbf{S2R-Synthesis}, a paired video reflection synthesis pipeline that performs physics-grounded augmentation in the structure space and uses a trained video renderer to generate controllable reflected videos.

    \item We introduce \textbf{S2R-Removal}, \textit{the first diffusion-based video reflection removal model} which removes reflections in one step and runs faster than non-diffusion baselines, together with \textbf{S2R-Bench}, \textit{the first benchmark for this task}, supporting both full-reference objective evaluation and real-world perceptual assessment.
\end{itemize}

Extensive experiments on S2R-Bench and multiple public image reflection removal benchmarks demonstrate state-of-the-art performance.
We further validate the effectiveness of S2R-Synthesis through data ablations.

\section{Related Work}
\label{sec:related}

\subsection{Single-Image and Video Reflection Removal}

Single-image reflection removal has progressed from handcrafted priors and multi-image constraints~\cite{levin2007user,guo2014robust} to deep models with bidirectional layer estimation and perceptual supervision~\cite{fan2017generic,yang2018seeing,zhang2018single,wan2018crrn}, and further to robustness-oriented designs using polarization, location awareness, dual-stream interaction, and in-the-wild modeling~\cite{lei2020polarized,dong2021location,hu2023single,zhao2025reversible,zhu2024revisiting}.
However, all such methods process frames independently and produce flickering when applied to video, and video reflection removal has only been touched by spatio-temporal optimization~\cite{nandoriya2017video} and user-guided decomposition~\cite{ahmed2021user}, leaving scalable learning-based priors unexplored.

\subsection{Reflection Data Synthesis and Benchmarks}

Paired reflection data are hard to collect, so existing image datasets rely on controlled capture, real-world collection, or synthetic composition under a linear layer formation model~\cite{zhang2018single,wei2019single,wan2022benchmarking,dong2021location,lei2020polarized,yang2025ntire,hu2026dereflection,hu2023single,wen2019single}, with recent variants adding non-linear alpha masks~\cite{wen2019single}, RAW-domain modeling~\cite{kee2025removing}, and physically based rendering~\cite{zakarin2025reflection}.
Video reflection additionally requires temporally coherent behavior and aligned ground truth, which no prior benchmark provides; our S2R-Bench fills this gap with the first paired-video benchmark and controllable synthesis pipeline.

\subsection{Diffusion Priors for Reflection Removal}

Diffusion priors have been applied to single-image dereflection via visual prompts~\cite{wang2024promptrr}, self-supervised separation~\cite{lu2025single}, one-step removal with diversified data~\cite{hu2026dereflection}, and latent-space prior modulation~\cite{li2025rectifying,xu2026fumo}, but remain confined to the image domain.
S2R-Removal extends diffusion-based dereflection to video for the first time.

\section{Physics-Grounded Reflection Simulation}
\label{sec:synthesis}

\begin{figure*}[t]
    \centering
    \includegraphics[width=\textwidth]{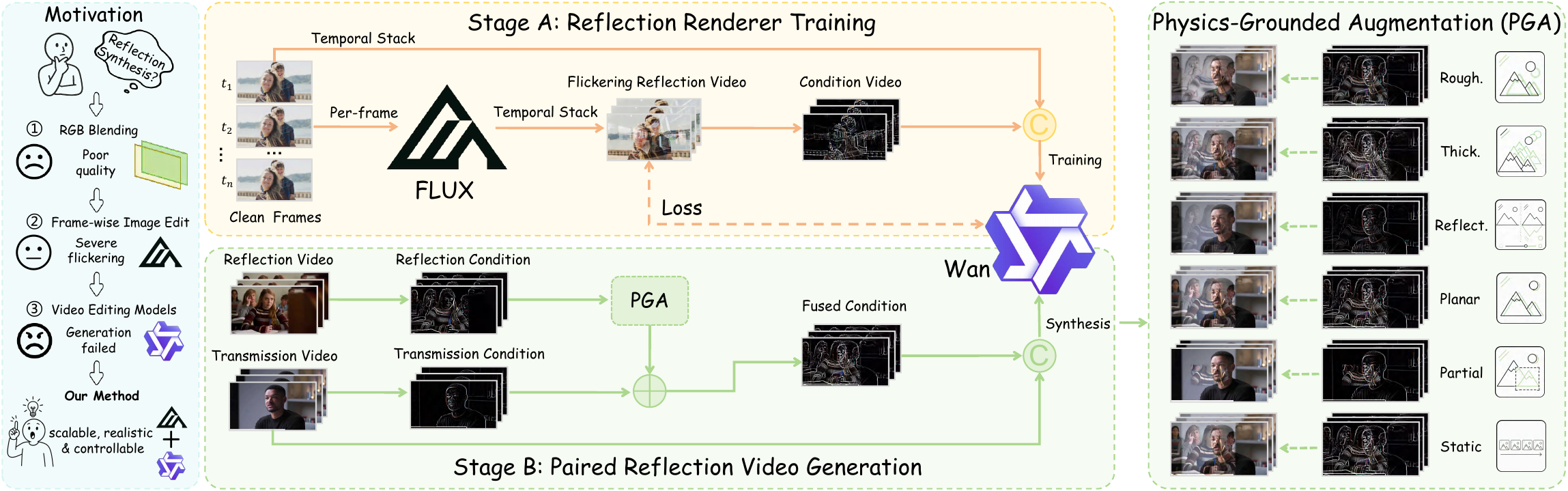}
    \caption{
    Overview of S2R-Synthesis.
    {Left:} limitations of existing strategies motivate our S2R-Synthesis.
    {Stage~A} trains a structure-guided reflection renderer from FLUX-generated pseudo reflection videos;
    {Stage~B} fuses two clean videos' conditions through Physics-Grounded Augmentation (PGA) and renders paired reflected videos.
    {Right:} representative PGA operations.
    }
    \label{fig:data_pipeline}
\end{figure*}

We build on the standard reflection formation model~\cite{hu2023single,levin2007user,wan2018crrn,wen2019single}, which represents the observed image as a linear mixture of transmission and reflection layers:
\begin{equation}
  I = \alpha_t T + \alpha_r R,
  \label{eq:reflection_model}
\end{equation}
where $T$ and $R$ denote the transmission and reflection layers, and $\alpha_t, \alpha_r$ control their relative contributions. Direct RGB-space blending under this model produces overly simplistic reflections and fails to capture glass-dependent effects~\cite{zakarin2025reflection} such as roughness-induced blur, thickness-induced ghosting, and reflectance variation.

\paragraph{Overview.}
We propose a two-stage pipeline that lifts reflection synthesis from RGB-space blending to \emph{structure-space conditioning}. As shown in~\Cref{fig:data_pipeline}, Stage~A trains a structure-guided reflection renderer $\mathcal{G}$, and Stage~B uses $\mathcal{G}$ together with a Physics-Grounded Augmentation (PGA) module to synthesize large-scale paired data:
\begin{equation}
  E_F = \mathcal{F}\!\bigl(E_T,\,\mathcal{A}(E_R;\theta_g)\bigr),
  \quad
  I   = \mathcal{G}(T, E_F),
  \label{eq:structure_rendering}
\end{equation}
where $E_T$ and $E_R$ are the lineart conditions~\cite{controlnet_aux} of the transmission and reflection source videos, $\mathcal{A}$ denotes the PGA module with glass control parameters $\theta_g$, $\mathcal{F}$ fuses the two linearts, and $\mathcal{G}$ renders a photorealistic reflection video from the clean video $T$ and the composed structural condition $E_F$.

\subsection{Video Reflection Renderer Training}

Given a collection of reflection-free videos selected by a VLM, we apply FLUX frame-by-frame with reflection-injection prompts to obtain pseudo reflection videos. Because FLUX operates independently per frame, the resulting clips may exhibit temporal flicker, so we use them only as pseudo supervision. Details are provided in~\Cref{sec_sup:data:synthesis} in the appendix.

Concretely, for each clean video $T$ and its pseudo reflection video $\tilde{M}$, we extract a lineart condition from $\tilde{M}$---capturing structural layout while suppressing unstable appearance details---and train the renderer $\mathcal{G}$ (built on Wan2.1~\cite{wan2025wan}) to reconstruct $\tilde{M}$ from $T$ and this lineart, supervised by a diffusion loss. We use a three-channel representation, consistent with the RGB conditioning format of the pretrained backbone, which yields more naturally colored reflections than a single-channel counterpart.

\subsection{Paired Reflection Video Generation}

Given a transmission video $T$ and an independent reflection source video $R$, we extract their lineart conditions $E_T$ and $E_R$. The reflection lineart $E_R$ is transformed by PGA, fused with $E_T$, and fed together with $T$ into $\mathcal{G}$ to produce a reflection video $I$ (Eq.~\eqref{eq:structure_rendering}).
The pair $(I, T)$ serves as training data for the reflection removal model.
Compared with RGB-space blending (Eq.~\eqref{eq:reflection_model}), compositing in lineart space and delegating appearance synthesis to the learned renderer yields more natural reflections with exact paired supervision.
Dataset construction details are provided in~\Cref{sec_sup:rem_data:paired} in the appendix.

\subsection{Physics-Grounded Augmentation}
\label{sec:PGA}

Real glass reflections vary with surface roughness, glass thickness, viewing angle, spatial coverage, and temporal behavior.
Rather than running full light-transport simulation, PGA instantiates the dominant visual effects of these factors as controllable operations on the reflection lineart $E_R$, governed by clip-level parameters $\theta_g=\{\sigma,\Delta,w,\gamma,\beta,p\}$ shared across frames for temporal coherence.

\paragraph{Augmentation Primitives.}
PGA composes six operations, each tied to a specific glass optical property (full derivations are in~\Cref{sec_sup:PGA} and parameter ranges are provided in~\Cref{tab:PGA} in the appendix):
\textit{Roughness} approximates the far-field angular spread of microfacet (GGX) scattering~\cite{walter2007microfacet,karis2013real} on rough glass as a Gaussian convolution of $E_R$ with width $\sigma$;
\textit{Thickness} reproduces multi-interface ghosting under paraxial Snell's law~\cite{born2013principles}, adding a shifted copy $w \cdot \mathcal{W}_{\Delta}(E_R)$ with offset $\Delta$ following the lateral displacement $\delta \approx d\,\theta_i(1-1/n)$;
\textit{Reflectance} instantiates the Fresnel decomposition~\cite{born2013principles} $I_\mathrm{refl} = F \cdot L_\mathrm{env} + L_\mathrm{amb}$ under spatially uniform incident angle as an affine modulation $\gamma \cdot E_R + \beta$;
\textit{Partial} captures partial glass coverage by masking $E_R$ with a spatial mask $p$;
\textit{Static} captures temporally stable reflections from stationary sources by freezing a randomly selected frame of $E_R$ across the clip;
\textit{Planar} serves as the default mode for flat-glass reflection, directly fusing $E_T$ and $E_R$.

\section{Diffusion-Based Video Dereflection}
\label{sec:removal}

\begin{figure*}[t]
    \centering
    \includegraphics[width=\textwidth]{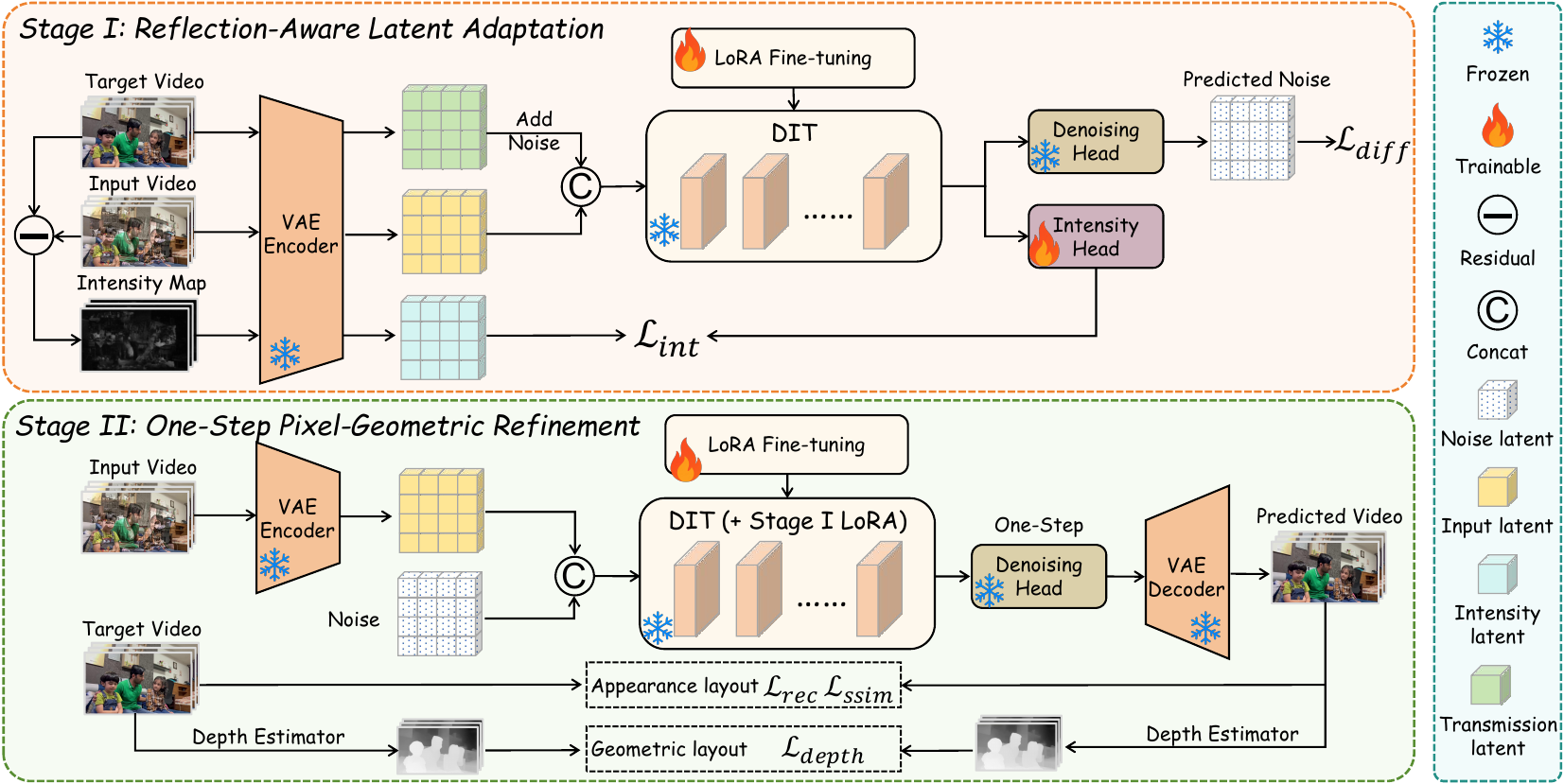}
    \caption{
Overview of S2R-Removal.
Stage~I learns reflection-aware latent adaptation via residual-derived intensity supervision from $(I,T)$;
Stage~II applies one-step pixel-geometric refinement with reconstruction, structural, and depth consistency losses.
}
    \label{fig:model_pipeline}
\end{figure*}

Given paired videos from our synthesis pipeline, we train a diffusion-based video reflection removal model.
We formulate the task as conditional video generation: conditioned on the reflected video $I$, the model predicts the clean transmission video $T$.
Built on a video inpainting paradigm using Wan2.1, our adaptation follows two principles: the model should be \textit{reflection-aware} (localizing and estimating reflection strength) and \textit{transmission-preserving} (removing reflections without altering scene content or geometry).
We address these with a two-stage training strategy, illustrated in~\Cref{fig:model_pipeline}.

\subsection{Stage I: Reflection-Aware Latent Adaptation}

The first stage adapts the pretrained video diffusion backbone to the reflected-to-clean mapping in latent space via LoRA fine-tuning.
Let \(z_T=\mathcal{E}(T)\) and \(z_I=\mathcal{E}(I)\) denote the latent representations of the clean transmission video and the reflected input video, respectively. We add noise to the clean target latent \(z_T\) and use the reflected input latent \(z_I\) as the condition.
Specifically, we sample \(t\sim\mathcal{U}(1,N)\) and \(\epsilon\sim\mathcal{N}(0,\mathbf{I})\), and obtain $z_t=\alpha_t z_T+\sigma_t\epsilon$.
Following the prediction parameterization of the pretrained backbone, the DiT predicts the diffusion target $u_t$ by the following diffusion loss:
\begin{equation}
\mathcal{L}_{\mathrm{diff}}
=
\mathbb{E}_{t,\epsilon}
\left[
\left\|
f_\theta(z_t,t,z_I)-u_t
\right\|_2^2
\right].
\label{eq:loss_diff}
\end{equation}

Our paired synthetic data provide a natural supervision signal for this.
We derive a continuous reflection-intensity map from the residual between the reflected input and the clean target, $M_{\mathrm{gt}} = |I - T|$, which encodes both the location and strength of reflection corruption.
To inject this signal into the backbone without disturbing its pretrained denoising behavior, we attach a lightweight, zero-initialized intensity head to the DiT features alongside the main denoising head.
The head is supervised by an $L_1$ loss in latent space:
\begin{equation}
  \mathcal{L}_{\mathrm{int}}
  = \bigl\|\hat{M} - \mathcal{E}(M_{\mathrm{gt}})\bigr\|_1,
  \label{eq:intensity_loss}
\end{equation}
where $\hat{M}$ is the predicted intensity map in the latent space and $\mathcal{E}$ is the VAE encoder.
The Stage~I objective combines both losses:
\begin{equation}
  \mathcal{L}_{\mathrm{stage1}}
  = \lambda_{\mathrm{diff}}\mathcal{L}_{\mathrm{diff}}
  + \lambda_{\mathrm{int}}\mathcal{L}_{\mathrm{int}},
  \label{eq:stage1_loss}
\end{equation}
with $\lambda_{\mathrm{diff}} = 1.0$ and $\lambda_{\mathrm{int}} = 0.1$.

Through this stage, the model acquires a strong prior on reflected-to-clean translation, and---as a by-product of the latent diffusion training---acquires a meaningful one-step denoising capability: a single forward pass already produces a rough but coherent clean estimate, consistent with the observation in~\cite{li2025rectifying}.
This emergent one-step capability is the key enabler of Stage~II.

\subsection{Stage II: One-Step Pixel-Geometric Refinement}

Latent diffusion training does not directly constrain the decoded output, and the model's generative freedom may alter texture or local structure in ways that are undesirable for reflection removal.
We therefore introduce a second stage that anchors the output to the clean target in both pixel and geometry space.

Crucially, this stage exploits the one-step denoising capability acquired in Stage I. Rather than running expensive multi-step sampling, we start from a noisy latent $z_\tau\sim\mathcal{N}(0,\mathbf{I})$ at the largest noise level $\tau$, and perform a single deterministic denoising update conditioned on the reflected input latent $z_I$:
\begin{equation}
\hat z_0
=
\mathcal{D}^{(1)}_\theta(z_\tau,\tau,z_I),
  \label{eq:onestep}
\end{equation}
where $\mathcal{D}^{(1)}_\theta$ denotes the one-step conversion from the model prediction to the clean latent estimate under the pretrained backbone's sampling parameterization. The predicted latent $\hat z_0$ is decoded by the frozen VAE decoder to obtain the pixel-space output $\hat T$.

\paragraph{Pixel-Space Losses.}
We apply an $L_1$ reconstruction loss and an SSIM loss to constrain color fidelity and local structural consistency:
\begin{equation}
  \mathcal{L}_{\mathrm{rec}}  = \|\hat{T} - T\|_1,
  \qquad
  \mathcal{L}_{\mathrm{ssim}} = 1 - \mathrm{SSIM}(\hat{T}, T).
  \label{eq:pixel_losses}
\end{equation}

\paragraph{Reflection-Aware Depth Consistency Loss.}
To further constrain scene geometry without over-penalizing style or texture, we introduce a depth consistency loss computed by a frozen LeReS~\cite{yin2021learning} depth estimator:
\begin{equation}
  D_{\hat{T}} = \Phi_{\mathrm{dep}}(\hat{T}), \qquad
  D_{T}       = \Phi_{\mathrm{dep}}(T).
  \label{eq:depth_maps}
\end{equation}
Gradients are propagated only through $D_{\hat{T}}$; $D_T$ is treated as a fixed target.

We derive a binary reflection mask $M = \mathbb{I}(|I - T| > \tau_m)$ with $\tau_m = 12$ to focus geometric supervision on reflection-corrupted regions, where hallucination and structural distortion are most likely:
\begin{equation}
  \mathcal{L}_{\mathrm{depth}}
  =
  \begin{cases}
    \dfrac{\sum M\,|D_{\hat{T}} - D_T|}{\sum M},
      & \sum M > 0, \\[8pt]
    \dfrac{1}{|\Omega|}\sum_{\Omega}|D_{\hat{T}} - D_T|,
      & \sum M = 0,
  \end{cases}
  \label{eq:depth_loss}
\end{equation}
where $\Omega$ denotes the full pixel domain.

The Stage~II objective is:
\begin{equation}
  \mathcal{L}_{\mathrm{stage2}}
  = \lambda_{\mathrm{rec}}\mathcal{L}_{\mathrm{rec}}
  + \lambda_{\mathrm{ssim}}\mathcal{L}_{\mathrm{ssim}}
  + \lambda_{\mathrm{dep}}\mathcal{L}_{\mathrm{depth}},
  \label{eq:stage2_loss}
\end{equation}
with $\lambda_{\mathrm{rec}} = 1.0$, $\lambda_{\mathrm{ssim}} = 0.2$,
and $\lambda_{\mathrm{dep}} = 0.5$.

The two stages are complementary: Stage~I teaches the model \emph{what to
remove} through reflection-intensity supervision in latent space, while
Stage~II teaches the model \emph{what to preserve} through pixel and
geometric constraints in image space---made efficient by the one-step
denoising capability that Stage~I instils.

\section{Experiments}
\label{experiments}

\subsection{Experimental Setup}

\paragraph{Implementation Details.}
Both the reflection synthesis and removal models are initialized from Wan2.1-Fun-v1.1-1.3B-Inp~\cite{wan2025wan}.
The synthesis model is trained for 8k steps on 21k paired samples, with pseudo reflection videos produced by FLUX.2-Klein-9B~\cite{flux-2-2025}.
The removal model follows our two-stage strategy, trained for 30k steps per stage on 38k pairs, where 80\% of the image samples are converted into video sequences by applying virtual camera motions (crop-to-video strategy~\cite{hu2026autoawg}).
All models are trained on 8 NVIDIA A100 GPUs. Further training details are provided in~\Cref{sec_sup:syn_data,sec_sup:rem_data} in the appendix.

\paragraph{Benchmark.}
We introduce \textbf{S2R-Bench}, the first benchmark dedicated to video reflection removal, comprising two complementary subsets.
\textbf{S2R-Ref} contains 60 paired videos constructed from the DRR dataset~\cite{hu2026dereflection}.
Each reflected/clean image pair is converted to a 10\,fps static video sequence with identical virtual camera motions applied to both, preserving pixel-level alignment.
\textbf{S2R-Real} contains 50 in-the-wild reflection videos collected from real captures and online sources, covering diverse glass materials, lighting conditions, reflection strengths, and camera motions, used for human perceptual evaluation and real-world generalization analysis.
Details are provided in~\Cref{sec_sup:bench} in the appendix.

We additionally evaluate cross-domain generalization on standard image reflection removal benchmarks: Real~\cite{zhang2018single}, Nature~\cite{dong2021location},  SIR$^2$~\cite{wan2022benchmarking}, and OpenRR-1k~\cite{yang2025ntire}.

\paragraph{Evaluation Metrics.}
On S2R-Ref and image benchmarks we report PSNR~\cite{hore2010image} and SSIM~\cite{wang2004image}.
For video evaluation we additionally report Temporal Consistency (TC)~\cite{zhang2024avid}.
For S2R-Real, 15 participants score each video from 0 to 3 on two aspects: \textit{reflection removal quality} and \textit{transmission preservation}; scores are averaged and linearly normalized for reporting. Details are provided in~\Cref{sec_sup:bench:eval} in the appendix.

\subsection{Comparison with State-of-the-Art}
\label{sec:sota_comparison}

\begin{table*}[t]
\centering
\caption{
Comparison with state-of-the-art dereflection methods on video and image benchmarks.
S2R-Ref reports full-reference metrics + TC; S2R-Real reports normalized human scores + TC; DAI is excluded from S2R-Ref to avoid DRR data leakage.
Per-frame inference time is measured on an 81-frame video; our one-step model is $\sim$1.67$\times$ faster than the next-best method.
}
\label{tab:sota_combined}
\setlength{\tabcolsep}{3.0pt}
\renewcommand{\arraystretch}{1.12}
\resizebox{\textwidth}{!}{
\begin{tabular}{lcccccc cccccccc c}
\toprule
\textbf{Method}
& \multicolumn{6}{c}{\textbf{Video Bench}}
& \multicolumn{8}{c}{\textbf{Image Bench}}
& \textbf{Efficiency} \\
\cmidrule(lr){2-7} \cmidrule(lr){8-15} \cmidrule(lr){16-16}
& \multicolumn{3}{c}{\textbf{S2R-Ref(60)}}
& \multicolumn{3}{c}{\textbf{S2R-Real(50)}}
& \multicolumn{2}{c}{\textbf{Real(20)}}
& \multicolumn{2}{c}{\textbf{SIR$^2$(454)}}
& \multicolumn{2}{c}{\textbf{Nature(20)}}
& \multicolumn{2}{c}{\textbf{Average}}
& \textbf{832 $\times$ 480} \\
\cmidrule(lr){2-4} \cmidrule(lr){5-7}
\cmidrule(lr){8-9} \cmidrule(lr){10-11} \cmidrule(lr){12-13} \cmidrule(lr){14-15} \cmidrule(lr){16-16}
& \textbf{PSNR$\uparrow$}
& \textbf{SSIM$\uparrow$}
& \textbf{TC$\uparrow$}
& \textbf{Removal$\uparrow$}
& \textbf{Preserv.$\uparrow$}
& \textbf{TC$\uparrow$}
& \textbf{PSNR$\uparrow$}
& \textbf{SSIM$\uparrow$}
& \textbf{PSNR$\uparrow$}
& \textbf{SSIM$\uparrow$}
& \textbf{PSNR$\uparrow$}
& \textbf{SSIM$\uparrow$}
& \textbf{PSNR$\uparrow$}
& \textbf{SSIM$\uparrow$}
& \textbf{ms/frame$\downarrow$} \\
\midrule
DSRNet~(ICCV'23)
& 23.78 & 0.857 & 0.9546
& 0.163 & 0.785 & {0.9790}
& 23.91 & 0.818
& 25.71 & 0.906
& 25.22 & 0.832
& 25.62 & 0.899
& 329.23 \\

DSIT~(NeurIPS'24)
& 24.06 & \textbf{0.870} & 0.9585
& 0.297 & 0.853 & 0.9781
& 25.22 & 0.836
& 26.43 & 0.911
& 26.77 & \underline{0.847}
& 26.40 & 0.905
& 241.05 \\

RDNet~(CVPR'25)
& 24.71 & \textbf{0.870} & 0.9564
& 0.175 & 0.867 & \underline{0.9797}
& 25.71 & \underline{0.850}
& 26.69 & 0.908
& 26.31 & 0.846
& 26.63 & 0.903
& \underline{145.13} \\

DAI~(AAAI'26)
& -- & -- & --
& 0.332 & \underline{0.871} & 0.9777
& 25.21 & 0.841
& 27.47 & 0.919
& 26.81 & 0.843
& 27.35 & 0.913
& 217.27 \\

GenSIRR~(CVPR'26)
& \underline{27.04} & 0.846 & \underline{0.9648}
& \underline{0.654} & 0.831 & 0.9747
& \underline{27.58} & \textbf{0.881}
& \underline{28.08} & \textbf{0.937}
& \underline{27.34} & 0.840
& \underline{28.03} & \textbf{0.931}
& 7214.94 \\

\midrule
Ours
& \textbf{28.84} & \underline{0.859} & \textbf{0.9740}
& \textbf{0.787} & \textbf{0.980} & \textbf{0.9827}
& \textbf{28.23} & \textbf{0.881}
& \textbf{28.83} & \underline{0.925}
& \textbf{27.89} & \textbf{0.883}
& \textbf{28.77} & \underline{0.921}
& \textbf{87.09} \\
\bottomrule
\end{tabular}
}
\end{table*}

\begin{figure*}
    \centering
    \includegraphics[width=\textwidth]{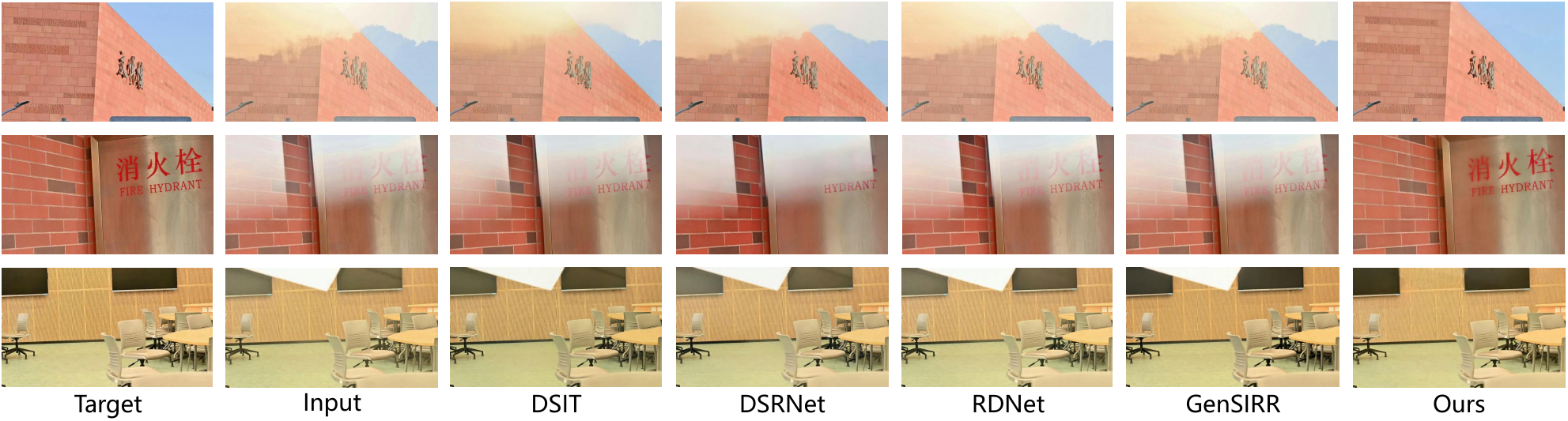}
    \caption{
    Qualitative comparison on controlled reflection videos with clean ground truth.
    Our method removes reflections more completely while preserving the transmission content.
    }
    \label{fig:qual_gt}
\end{figure*}

We compare against recent image reflection removal methods --- DSRNet~\cite{hu2023single}, DSIT~\cite{hu2024single}, RDNet~\cite{zhao2025reversible}, DAI~\cite{hu2026dereflection}, and GenSIRR~\cite{li2025rectifying} --- applied frame-by-frame as video baselines.
DAI is excluded from S2R-Ref because S2R-Ref is constructed from DAI's training data.
Existing video reflection removal methods~\cite{nandoriya2017video,ahmed2021user} release no code, so we compare against them only qualitatively on their test cases (see~\Cref{sec_sup:additional_qual}).

\paragraph{Quantitative results.}
As shown in~\Cref{tab:sota_combined}, our method achieves the best PSNR and TC on S2R-Ref and the highest human scores on S2R-Real for both reflection removal and transmission preservation.
Despite being trained for video dereflection, it also delivers competitive or superior performance on image benchmarks, indicating strong cross-domain generalization.
Beyond accuracy, our one-step design yields a substantial efficiency advantage: S2R-Removal runs at 87.09\,ms/frame, $\sim$1.67$\times$ faster than the next-best method RDNet.

\begin{figure*}
    \centering
    \includegraphics[width=\textwidth]{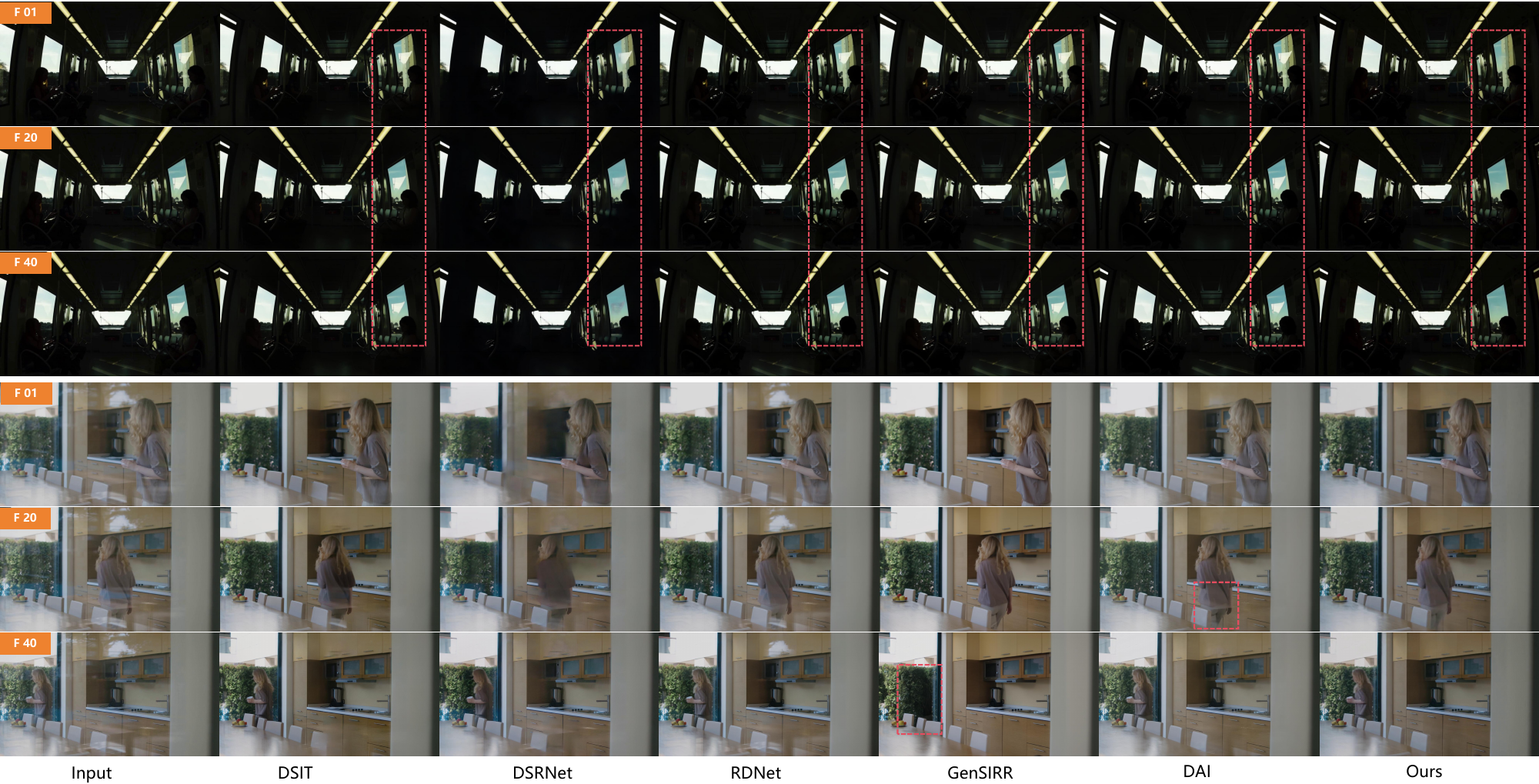}
    \caption{
    Qualitative comparison with SOTA reflection removal methods on videos (1st/20th/40th frame).
    Existing methods leave residues or temporally inconsistent removal, while our S2R-Removal produces coherent reflection removal.
    }
    \label{fig:sup_frame_image_methods}
\end{figure*}

\paragraph{Qualitative results.}
As shown in~\Cref{fig:qual_gt,fig:sup_frame_image_methods}, image-based methods often leave reflection residues, introduce temporal flickering, or shift the global appearance of the scene, whereas our method suppresses the reflected layer more accurately while preserving the original tone and structure of the transmission content.
Further qualitative results are provided in~\Cref{sec_sup:additional_qual}.

\begin{table}[t]
\centering
\caption{
Effectiveness of our synthesis pipeline.
RDNet is trained on OpenRR-1k\_train with different data configurations and evaluated on OpenRR-1k\_test.
Legend: Trad.~\cite{hu2023single}; P/St/Pa/Re/Th/Ro = ours with Planar/Static/Partial/Reflectance/Thickness/Roughness; $\Delta$ is relative to the first row of each panel.
}
\label{tab:synthesis_effectiveness}
\renewcommand{\arraystretch}{1.08}
\setlength{\tabcolsep}{3.0pt}
\begin{tabular}{llcccc}
\toprule
\textbf{ID} & \textbf{Setting} & \textbf{PSNR$\uparrow$} & \textbf{$\Delta$PSNR} & \textbf{SSIM$\uparrow$} & \textbf{$\Delta$SSIM} \\
\midrule
\multicolumn{6}{l}{\textit{A: Synthetic data source}} \\
\midrule
A1 & Base only & 28.05 & +0.00 & 0.9465 & +0.0000 \\
A2 & + Trad. & 31.98 & +3.93 & 0.9619 & +0.0154 \\
A3 & + Ours(P) & 32.35 & +4.31 & 0.9643 & +0.0178 \\
A4 & + Trad. + Ours(P) & \textbf{33.26} & \textbf{+5.21} & \textbf{0.9679} & \textbf{+0.0214} \\
\midrule
\multicolumn{6}{l}{\textit{B: PGA variants}} \\
\midrule
B1 & P & 32.35 & +0.00 & 0.9643 & +0.0000 \\
B2 & P + St & 32.81 & +0.46 & 0.9670 & +0.0027 \\
B3 & P + St + Pa & 33.62 & +1.27 & 0.9691 & +0.0048 \\
B4 & P + St + Pa + Re & 33.58 & +1.23 & 0.9695 & +0.0052 \\
B5 & P + St + Pa + Re + Th & 33.70 & +1.35 & 0.9697 & +0.0054 \\
B6 & Full PGA & \textbf{34.13} & \textbf{+1.78} & \textbf{0.9704} & \textbf{+0.0061} \\
\bottomrule
\end{tabular}

\end{table}

\subsection{Effectiveness of S2R-Synthesis}
\label{sec:synthesis_exp}

To isolate the effect of data generation, we fix the backbone to the state-of-the-art image reflection removal model RDNet~\cite{zhao2025reversible} and vary only the training data.
All models are trained on OpenRR-1k trainset and evaluated on OpenRR-1k testset.

\Cref{tab:synthesis_effectiveness} reports results across two panels.
\textbf{Panel A} compares data sources. Adding traditional layer-blending synthesis~\cite{hu2023single} to the baseline already yields a clear gain, confirming the value of synthetic supervision.
Replacing it with our diffusion-rendered planar data further improves PSNR from 31.98\,dB to 32.35\,dB and SSIM from 0.9619 to 0.9643, and combining both sources achieves the best result in Panel~A, suggesting complementarity.

\textbf{Panel B} evaluates our Physics-Grounded Augmentation (PGA).
Adding each type of augmentations progressively improves overall performance.
Full PGA reaches 34.13\,dB PSNR and 0.9704 SSIM, a gain of +1.78\,dB and +0.0061 SSIM over the planar-only setting, validating that physically motivated controls substantially improve the diversity and training value of synthetic data.

\subsection{Ablation Study}
\label{sec:ablation_s2rdiff}

\begin{table}[t]
\centering
\caption{
Ablation study of S2R-Removal on S2R-Ref.
We progressively enable the Stage-I latent adaptation losses and the Stage-II pixel-geometric refinement losses.
}
\label{tab:s2rdiff_ablation}

\renewcommand{\arraystretch}{1.08}
\setlength{\tabcolsep}{2.4pt}

\begin{tabular*}{\columnwidth}{@{\extracolsep{\fill}}lccccccc}
\toprule
\multirow{2}{*}{\textbf{ID}}
& \multicolumn{2}{c}{\textbf{Stage-I}}
& \multicolumn{3}{c}{\textbf{Stage-II}}
& \multirow{2}{*}{\textbf{PSNR$\uparrow$}}
& \multirow{2}{*}{\textbf{SSIM$\uparrow$}} \\
\cmidrule(lr){2-3} \cmidrule(lr){4-6}
& $\mathcal{L}_{\mathrm{diff}}$
& $\mathcal{L}_{\mathrm{int}}$
& $\mathcal{L}_{\mathrm{rec}}$
& $\mathcal{L}_{\mathrm{ssim}}$
& $\mathcal{L}_{\mathrm{depth}}$
&  &  \\
\midrule
1 & \checkmark & --         & --         & --         & --         & 27.00 & 0.8348 \\
2 & \checkmark & \checkmark & --         & --         & --         & 27.71 & 0.8430 \\
3 & \checkmark & \checkmark & \checkmark & --         & --         & 28.21 & 0.8513 \\
4 & \checkmark & \checkmark & \checkmark & \checkmark & --         & 28.55 & 0.8566 \\
5 & \checkmark & \checkmark & \checkmark & \checkmark & \checkmark & \textbf{28.84} & \textbf{0.8594} \\
\bottomrule
\end{tabular*}

\end{table}

We ablate the key training losses of S2R-Removal on S2R-Ref, as summarized in~\Cref{tab:s2rdiff_ablation}.
Starting from the diffusion-only baseline, the residual-derived intensity supervision improves PSNR from 27.00\,dB to 27.71\,dB and SSIM from 0.8348 to 0.8430, showing that explicit reflection-intensity guidance benefits reflection-aware latent adaptation.
Afterwards, the Stage-II pixel-geometric refinement losses provide consistent gains: $\mathcal{L}_{\mathrm{rec}}$ improves reconstruction fidelity, $\mathcal{L}_{\mathrm{ssim}}$ enhances structural consistency, and $\mathcal{L}_{\mathrm{depth}}$ further regularizes scene geometry.
The full objective achieves the best result, with 28.84\,dB PSNR and 0.8594 SSIM.
Qualitative effects of two stages are provided in~\Cref{sec_sup:attmap} in the appendix.

\section{Conclusion}
\label{sec:conclusion}

We presented a closed-loop framework for video reflection removal, spanning physics-grounded paired video synthesis, reflection-aware diffusion-based dereflection, and benchmark evaluation. 
Our synthesis pipeline performs structure-space reflection composition with Physics-Grounded Augmentation and a learned video renderer, enabling large-scale paired training data generation. 
Based on this data, S2R-Removal adapts a pretrained video diffusion prior with reflection-intensity supervision and pixel-geometric refinement for temporally coherent reflection removal. 
Together with S2R-Bench, our framework provides a complete foundation for training and evaluating video dereflection methods.

\section*{Acknowledgements}

This work uses the FLUX.2-klein-9B model, licensed under FLUX Non-Commercial License. The Ditto-1M, UltraVideo, OpenRR-1k and OpenRR-5k datasets licensed under CC BY-NC-SA 4.0. The SIR2 dataset licensed for non-commercial purposes. The authors confirm that all uses of the above resources are strictly for academic research purposes and not for any commercial application.

\bibliography{ref}

\clearpage
\setcounter{page}{1}
\appendix
\setcounter{tocdepth}{2}

\section*{Appendix}

\noindent This appendix is organized as follows:
\begin{itemize}
    \item \textbf{A1. Reflection Synthesis Training Data}: VLM-filtered FLUX-based pseudo reflection video generation and the resulting paired data composition.
    \item \textbf{A2. Reflection Removal Training Data}: Paired video synthesis from HumanVid and UltraVideo, with the full training data composition.
    \item \textbf{A3. Physics-Grounded Augmentation}: Detailed derivations and operations of the six PGA primitives, with a trend comparison against a UE5-based pipeline.
    \item \textbf{A4. S2R-Bench and Human Evaluation}: Construction of the S2R-Ref and S2R-Real subsets and the human perceptual scoring criteria.
    \item \textbf{A5. Additional Qualitative Comparisons}: Real-world comparison, comparison with a general-purpose video editing model, and comparison with prior video dereflection methods.
    \item \textbf{A6. Component Visualization}: Cross-attention response of the reflection-intensity head in Stage~I, and the qualitative effect of Stage~II pixel-geometric refinement.
    \item \textbf{A7. Application to Single Images}: Zoom-based video inference for applying the video model to single-image reflection removal, with cross-domain qualitative evaluation on image benchmarks.
    \item \textbf{A8. Downstream Benefits of Dereflection}: Driving area segmentation and vehicle detection on BDD100K before and after dereflection.
    \item \textbf{A9. Limitations}: Discussion of method limitations on multi-layer reflections and coupled camera-motion geometry.
\end{itemize}

\section{Reflection Synthesis Training Data}
\label{sec_sup:syn_data}

\subsection{FLUX-Based Pseudo Reflection Video Generation}
\label{sec_sup:data:synthesis}

To construct paired training data for video reflection synthesis, we first filter reflection-free videos from the publicly available Ditto-1M dataset~\cite{bai2025scaling} using a Vision-Language Model (VLM), specifically Qwen3-VL-8B-Instruct~\cite{bai2025qwen3}. Each video is evaluated, and only those containing no glass or eyeglass reflections are retained. We design the following prompt for the VLM:

\begin{quote}
\textit{``You are a visual AI assistant. Analyze the content of the video and answer precisely. Task: Determine whether there is any glass reflection or eyeglass reflection visible in the video frames. If there is any frame containing glass or eyeglass reflections, output only: True. If there are no reflections at all in all frames, output only: False. Do NOT provide explanations, descriptions, or extra text. Only output True or False.''}
\end{quote}

For each filtered video, we apply image-to-image synthesis using FLUX.2-Kl\-ein-9B on a per-frame basis to add synthetic glass reflections. To introduce diversity in reflection intensity and appearance, we define a set of three prompts, and a single prompt is applied consistently across all frames of a given video:

\begin{itemize}[leftmargin=2em]
    \item \textbf{Subtle reflection:} \textit{``Preserve the original image unchanged. Only add a subtle glass reflection overlay on top of the scene. The reflection should be faint, transparent, and natural, with soft highlights and slight glare. Do not alter the structure, objects, colors, or composition of the original image.''}

    \item \textbf{Strong reflection:} \textit{``Preserve the original image unchanged. Only add strong, multi-layered glass reflections over the scene. The reflections should be vivid, highly visible, and complex, with overlapping glare, pronounced highlights, and multiple reflection sources. Do not alter the structure, objects, colors, or composition of the original image.''}

    \item \textbf{Colored reflection:} \textit{``Preserve the original image completely unchanged. Only overlay a colored glass reflection effect onto the scene. The reflections should be vivid, clear, and complex, with clearly visible light reflection sources. Do not alter the structure, objects, colors, or composition of the original image.''}
\end{itemize}

This stage yields \textbf{11,132 paired video clips} with inter-frame flickering reflections, where each pair consists of the original reflection-free frame and its synthesized reflection counterpart.

\subsection{Training Data Summary}
\label{sec_sup:data:summary}

Using the pipeline described above, we generate 11,132 paired video clips from Ditto-1M~\cite{bai2025scaling} as training data for S2R-Synthesis model. To further leverage existing image resources (incl. OpenRR-1k~\cite{yang2025ntire}, 
OpenRR-5k~\cite{cai2025openrr}, RRW~\cite{zhu2024revisiting}), we apply a crop-to-video strategy~\cite{hu2026autoawg} to convert existing image reflection datasets into short video clips, serving as supplementary training data. The full training dataset is summarized in~\Cref{tab:data}.

\begin{table}[t]
\centering
\caption{Composition of the reflection synthesis training dataset.}
\label{tab:data}
\begin{tabular}{llr}
\toprule
\textbf{Source Modality} & \textbf{Source} & \textbf{Samples} \\
\midrule
\multirow{5}{*}{Image} & OpenRR-1k   & 800   \\
                       & OpenRR-5k   & 5,000 \\
                       & RRW         & 3,000 \\
\midrule
{Video} & Ditto-1M (FLUX frame-wise synthetic) & 11,132 \\
\midrule
\multicolumn{2}{l}{\textbf{Total}} & \textbf{19,932} \\

\bottomrule
\end{tabular}

\end{table}

\section{Reflection Removal Training Data}
\label{sec_sup:rem_data}

\subsection{Physics-Grounded Paired Video Synthesis}
\label{sec_sup:rem_data:paired}

We first leverage the trained reflection synthesis model to generate paired reflection videos without inter-frame flickering.
Specifically, we filter reflection-free videos from the publicly available HumanVid~\cite{wang2024humanvid} and UltraVideo~\cite{xue2025ultravideo} datasets using a VLM, and synthesize 20,031 and 5,768 paired reflection videos respectively. The PGA parameters used during synthesis are summarized in~\Cref{tab:pga_params}.
In addition, we synthesize 2,997 paired reflection videos using the traditional synthesis method.

\subsection{Dataset Composition}
\label{sec_sup:rem_data:composition}
The training set for the reflection removal model contains \textbf{38,826 samples} in total, comprising both image and video data, as summarized in~\Cref{tab:removal_data}.
The image subset consists of 10,030 samples, identical to those used in reflection synthesis model training. During training, 80\% of the image samples are dynamically converted into video sequences by applying virtual camera motions (crop-to-video strategy~\cite{hu2026autoawg}).
The video subset consists of 28,796 samples. Of these, 25,799 are synthesized from HumanVid (20,031) and UltraVideo (5,768) using our physics-grounded pipeline, and 2,997 are synthesized from UltraVideo using a traditional RGB-space blending baseline~\cite{hu2023single}.

\begin{table}[t]
\centering
\caption{PGA augmentation parameters used for video reflection synthesis.}
\label{tab:pga_params}
\begin{tabular}{llll}
\toprule
\textbf{Augmentation} & \textbf{Parameter} & \textbf{Range} & \textbf{Probability} \\
\midrule
Roughness     & $\sigma$          & $[5.0, 15.0]$    & $0.5$ \\
\midrule
\multirow{3}{*}{Thickness}   & shift $\Delta_x$  & $[-50, 50]$\,  & \multirow{3}{*}{$0.5$} \\
                    & shift $\Delta_y$  & $[-50, 50]$\,  & \\
                    & weight $w$        & $[0.3, 0.7]$     & \\
\midrule
\multirow{2}{*}{Reflectance} & scale $\gamma$    & $[0.5, 2.5]$     & \multirow{2}{*}{$0.5$} \\
                    & bias $\beta$      & $[0, 0.15]$      & \\
\midrule
Partial      & num boxes         & $[3, 9]$         & $0.7$ \\
\midrule
Static   & —                 & —                & $0.1$ \\
\midrule
Planar & —                 & —                & $0.1$ \\
\bottomrule
\end{tabular}
\end{table}

\begin{table}[t]
\centering
\caption{Composition of the reflection removal training dataset.}
\label{tab:removal_data}
\begin{tabular}{llr}
\toprule
\textbf{Modality} & \textbf{Source} & \textbf{Samples} \\
\midrule
\multirow{5}{*}{Image} & OpenRR-1k   & 800   \\
                       & OpenRR-5k   & 5,000 \\
                       & RRW         & 3,000 \\
\midrule
\multirow{4}{*}{Video} & HumanVid (our synthesis)   & 20,031 \\
                       & UltraVideo (our synthesis) & 5,768  \\
                       & UltraVideo (traditional synthesis)      & 2,997  \\
\midrule
\multicolumn{2}{l}{\textbf{Total}} & \textbf{37,596} \\

\bottomrule
\end{tabular}
\end{table}

\section{Physics-Grounded Augmentation}
\label{sec_sup:PGA}
Given a transmission video $T$ and a reflection source video $R$, we first extract their respective lineart conditions $E_T$ and $E_R$, then apply a series of physics-grounded augmentations to the reflection lineart $E_R$, and finally fuse the augmented reflection condition with the transmission condition to obtain the composed structural condition. \Cref{tab:PGA} summarizes the correspondence between glass effects, physical cues, and our control-space operations.
The detailed augmentation parameters are summarized in~\Cref{tab:pga_params}.

\begin{table*}[t]
\centering
\caption{Physics-Grounded Augmentation (PGA) control variables and their physical interpretations.}
\label{tab:PGA}
\setlength{\tabcolsep}{6pt}
\resizebox{\linewidth}{!}{
\begin{tabular}{llll}
\toprule
\textbf{Augmentation} & \textbf{Glass effect} & \textbf{Physical cue} & \textbf{Structure-space operation} \\
\midrule
Planar & Flat glass reflection & Planar interface & Direct fusion \\
Roughness & Surface scattering & Microfacet roughness & Gaussian blur $(\sigma)$ \\
Thickness & Ghosting / secondary reflection & Refractive displacement & Translation / secondary offset $(\Delta,w)$ \\
Reflectance & Reflection strength variation & Fresnel reflectance & Intensity modulation $(\gamma,\beta)$ \\
Partial & Local glass coverage / occlusion & Spatial coverage & Random spatial mask $(p)$ \\
Static & Temporally stable reflection & Stationary reflection source & Frozen reflection frame \\
\bottomrule
\end{tabular}
}
\end{table*}

\subsection{Per-Primitive Physical Derivation}
\paragraph{Roughness Augmentation.}
Under the microfacet model~\cite{walter2007microfacet}, a rough glass surface distributes reflected rays over a wide angular lobe according to the GGX normal distribution function; in the far-field regime this angular spread is equivalent to convolving the ideal mirror reflection with an isotropic kernel whose width grows with roughness~\cite{karis2013real}.
We reproduce this effect by applying a channel-wise depthwise separable Gaussian convolution to the reflection lineart.
The standard deviation $\sigma$ is randomly sampled from $[5.0, 15.0]$, and the kernel size is automatically determined as $k = 2\lfloor 3\sigma \rfloor + 1$ to ensure effective coverage of the Gaussian distribution within the $\pm 3\sigma$ range.
The 2D Gaussian kernel is constructed as the outer product of a 1D kernel:
\begin{equation}
    g_i = \exp\!\left(-\frac{i^2}{2\sigma^2}\right), \quad
    \mathbf{K} = \frac{\mathbf{g}\mathbf{g}^\top}{\|\mathbf{g}\|_1^2},
\end{equation}
and applied via frame-wise spatial convolution over the video. 

\paragraph{Thickness Augmentation.}
Light incident on thick or multi-layer glass undergoes partial reflection at each interface, producing multiple reflections with lateral offsets. Under the paraxial approximation of Snell's Law~\cite{born2013principles}, the lateral displacement between two such reflections is proportional to glass thickness $d$, $\delta \approx d\,\theta_i(1 - 1/n)$, where $\theta_i$ is the incident angle and $n$ the refractive index. Motivated by this multi-interface displacement, we apply a spatial translation to the reflection lineart to simulate ghosting artifacts caused by refraction in thick glass.
The horizontal offset $\Delta_x$ and vertical offset $\Delta_y$ are independently sampled from $[-50, 50]$ pixels, and a translated lineart is generated via crop-and-pad operations.
The blending weight $w$ is sampled from $[0.3, 0.7]$, and the augmentation is formulated as:
\begin{equation}
    E_R^{thick} = E_R + w \cdot \mathcal{W}_{(\Delta_x, \Delta_y)}(E_R),
\end{equation}
where $\mathcal{W}_{(\Delta_x,\Delta_y)}$ denotes the pixel-wise translation operation with zero-padding at the boundaries.

\paragraph{Reflectance Augmentation.}
Following radiometric image formation, the observed reflection component can be decomposed into a view-dependent specular term and a diffuse ambient term, $I_{\mathrm{refl}} = F \cdot L_{\mathrm{env}} + L_{\mathrm{amb}}$, where $L_{\mathrm{env}}$ is the directional environmental radiance, $F$ is the Fresnel reflectance~\cite{born2013principles}, and $L_{\mathrm{amb}}$ accounts for the residual ambient illumination that contributes a spatially uniform brightness offset to the reflection layer. Under the assumption of spatially uniform incident angle, $F$ reduces to a scalar, which motivates an element-wise linear intensity modulation applied to the reflection lineart to simulate the viewing-angle-dependent reflection strength variation.
The scale factor $\gamma$ is sampled from $[0.5, 2.5]$ and the bias $\beta$ is sampled from $[0, 0.15]$. The augmentation is formulated as:
\begin{equation}
    E_R^{refl} = \gamma \cdot E_R + \beta.
\end{equation}
$\gamma < 1$ simulates weak reflections from transparent glass, $\gamma > 1$ simulates strong specular reflections, and $\beta$ introduces a brightness bias to account for ambient illumination.

\paragraph{Partial Coverage Augmentation.}
A random binary spatial mask is applied to the reflection lineart to simulate partial glass coverage or spatially varying reflectance.
$N_b \in [3, 9]$ rectangular regions are randomly generated, with the top-left and bottom-right coordinates of each rectangle independently sampled from the normalized image coordinate space $[0, 1]$.
The union of all rectangles forms the final mask $P \in \{0, 1\}^{H \times W}$, and the augmentation is formulated as $E_R^{crop} = E_R \odot P$, zeroing out the reflection structure outside the masked regions.

\paragraph{Static Reflection Augmentation.}
A randomly selected frame $f_r$ from the reflection source video is repeated along the temporal dimension across the entire clip, such that $E_R^{static}[f] = E_R[f_r]$ for all $f \in \{1, \ldots, F\}$, simulating a reflection source that remains stationary relative to the camera and ensuring complete temporal stability of the reflection layer.

\subsection{Condition Fusion and Physical Validation}
The augmented reflection lineart and the transmission lineart are summed element-wise and clipped to $[0, 1]$, yielding the final composed three-channel structural condition:
\begin{equation}
    E_F = \text{clip}\left(E_T + \mathcal{A}(E_R;\, \theta_g),\, 0,\, 1\right),
\end{equation}
where $\mathcal{A}$ denotes the composition of the above augmentation operations and $\theta_g$ denotes the randomly sampled augmentation parameters.
All parameters are sampled at the clip level and shared across frames to ensure temporal coherence.

\begin{figure*}[t]
    \centering
    \includegraphics[width=1.0\textwidth]{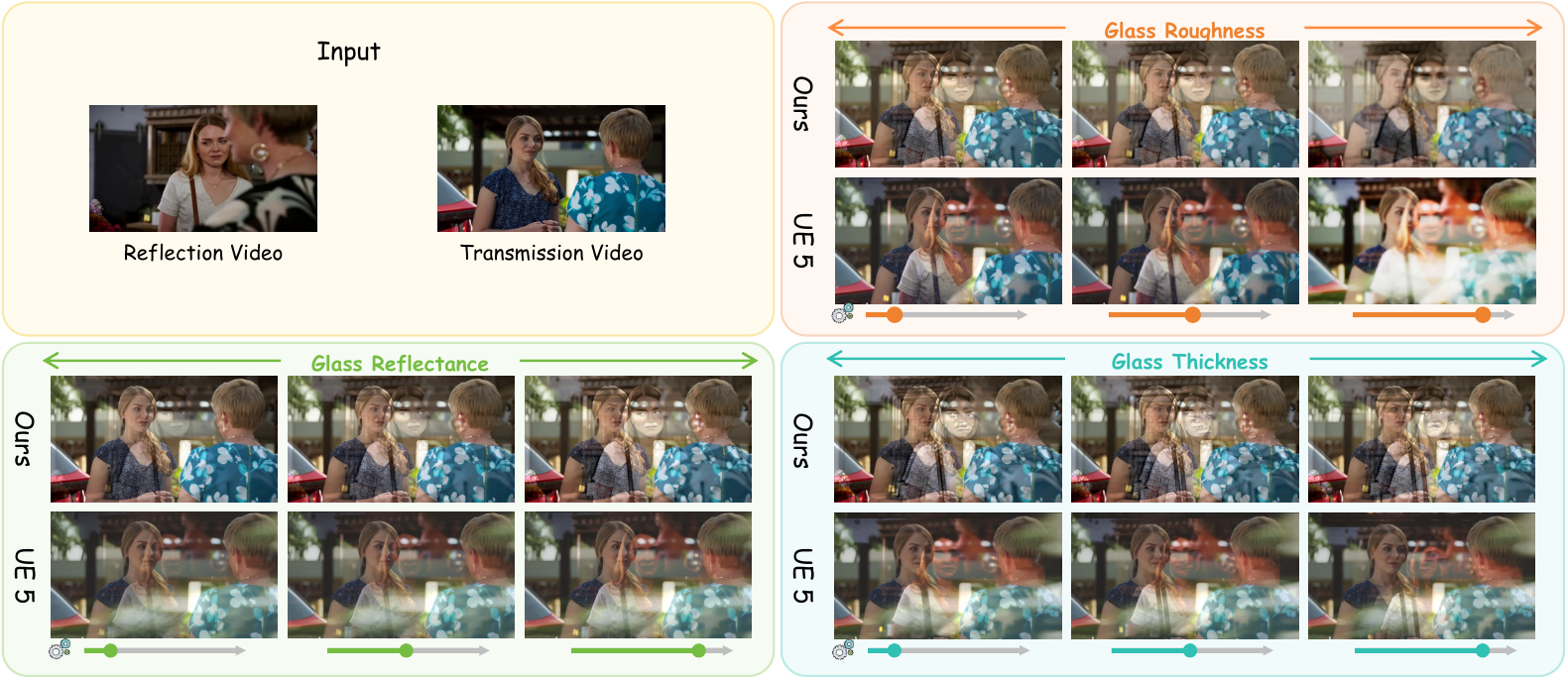}
    \caption{
    Comparison of reflection-synthesis trends between S2R-Synthesis (with PGA) and a UE5-based pipeline under controlled variation of three glass properties (roughness, reflectance, thickness).
    }
    \label{fig:synthesis_vs_ue5}
\end{figure*}

\Cref{fig:synthesis_vs_ue5} compares S2R-Synthesis with a UE5-based reflection synthesis pipeline under controlled variation of three glass properties: roughness, reflectance, and thickness.
Since the two pipelines operate under different rendering formulations, their numerical parameters are not directly comparable; we therefore focus on the qualitative trend each produces as a single property is varied.
In both pipelines, increasing roughness broadens the reflection blur, larger reflectance strengthens the overlay, and larger thickness amplifies the ghosting offset.
S2R-Synthesis reproduces these physically expected trends through lightweight structure-space controls, without requiring a heavyweight rendering engine.

\begin{table}[t]
\centering
\caption{Human perceptual evaluation scoring criteria.}
\label{tab:human_eval}
\begin{tabular}{llc}
\toprule
\textbf{Dimension} & \textbf{Description} & \textbf{Score} \\
\midrule
\multirow{4}{*}{Removal Quality}      & Complete removal                        & 3 \\
                                       & Major reflection components removed     & 2 \\
                                       & Slight removal                          & 1 \\
                                       & Failure                                 & 0 \\
\midrule
\multirow{4}{*}{Region Preservation}  & No visible degradation                  & 3 \\
                                       & Minor degradation                       & 2 \\
                                       & Noticeable degradation                  & 1 \\
                                       & Severe degradation                      & 0 \\
\bottomrule
\end{tabular}
\end{table}

\section{S2R-Bench and Human Evaluation}
\label{sec_sup:bench}

\subsection{S2R-Bench Construction}
\label{sec_sup:bench:con}
S2R-Bench consists of two complementary subsets: a full-reference subset \textbf{S2R-Ref} for quantitative evaluation and a no-reference subset \textbf{S2R-Real} for human perceptual evaluation.
\paragraph{S2R-Ref.}
S2R-Ref is constructed from the training data of the DRR dataset~\cite{hu2026dereflection}.
We first convert the reflection-contaminated frames and their corresponding clean frames into static video sequences at 10 fps, yielding 434 candidate paired video sequences in total.
However, since the paired frames in DRR are obtained via controlled capture, some sequences exhibit synthesis artifacts or inter-frame flickering that would compromise evaluation reliability.
We therefore conduct manual inspection and retain only sequences with visually clean and temporally consistent reflection appearance, resulting in 60 high-quality paired video sequences.

To simulate realistic camera motion, we apply identical virtual camera motions to both videos in each pair using the Ken Burns effect, which simulates lens movement through cropping and scaling.
For each pair, we randomly select one of two motion modes: Pan or Zoom.

In \textit{Pan} mode, a scale factor $s \in [1.5, 2.0]$ is first sampled uniformly at random.
A start crop position $(x_0, y_0)$ and an end crop position $(x_1, y_1)$ are then sampled within the scaled image space, subject to the constraint that their relative displacement exceeds 20\% of the image width or height, ensuring sufficient motion magnitude.

In \textit{Zoom} mode, the crop center is fixed while the scale factor varies over time.
The start and end scale factors are sampled from $[1.2, 1.5]$ and $[1.7, 2.0]$ respectively (or vice versa), simulating a zoom-in or zoom-out effect.

For each frame $i$, the crop position and scale are linearly interpolated according to the temporal progress $t = i / (N - 1) \in [0, 1]$:
\begin{equation}
\left\{
\begin{aligned}
x_t &= x_0(1-t) + x_1 t, \\
y_t &= y_0(1-t) + y_1 t, \\
s_t &= s_0(1-t) + s_1 t.
\end{aligned}
\right.
\end{equation}

The resulting crop region is then resized back to the original resolution.
Since both videos share the same motion parameters, the reflected video and the clean ground truth undergo identical frame-wise transformations, strictly preserving pixel-level alignment while introducing realistic camera dynamics.
This design supports full-reference quantitative evaluation with metrics including PSNR and SSIM.

\paragraph{S2R-Real.}
S2R-Real contains 50 in-the-wild reflection videos, each consisting of 81 frames, collected from real-world captures and online sources, covering diverse glass materials, lighting conditions, reflection strengths, and camera motion patterns.
Among them, 19 samples are dynamic videos with noticeable scene or camera motion, while the remaining 31 samples are static videos with relatively stable content.
Since ground-truth clean videos are unavailable in real-world scenarios, this subset is used for human perceptual evaluation.

\begin{figure*}[t]
    \centering
    \includegraphics[width=\textwidth]{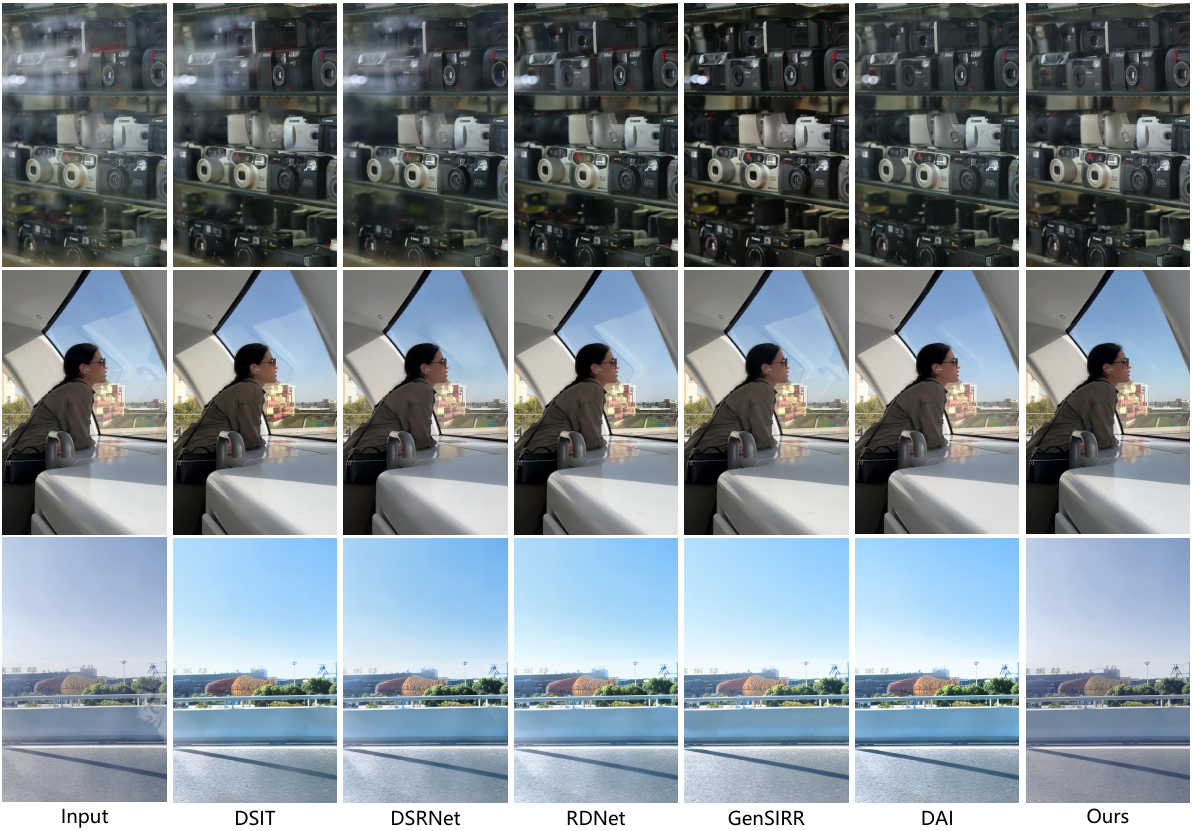}
    \caption{Qualitative comparison on real-world reflection videos without ground truth.}
    \label{fig:qual_real}
\end{figure*}

\subsection{Human Perceptual Evaluation}
\label{sec_sup:bench:eval}

The human perceptual evaluation criteria are summarized in~\Cref{tab:human_eval}.
For reflection removal quality, scores range from 0 to 3, where 3 denotes complete removal, 2 denotes removal of major reflection components, 1 denotes slight removal, and 0 denotes failure.
For non-reflection region preservation, scores also range from 0 to 3, where 3 denotes no visible degradation, 2 denotes minor degradation, 1 denotes noticeable degradation, and 0 denotes severe degradation.
We report the normalized removal and preservation scores separately, where higher values indicate better reflection suppression and stronger transmission preservation.

\begin{figure*}[t]
    \centering
    \includegraphics[width=1.0\textwidth]{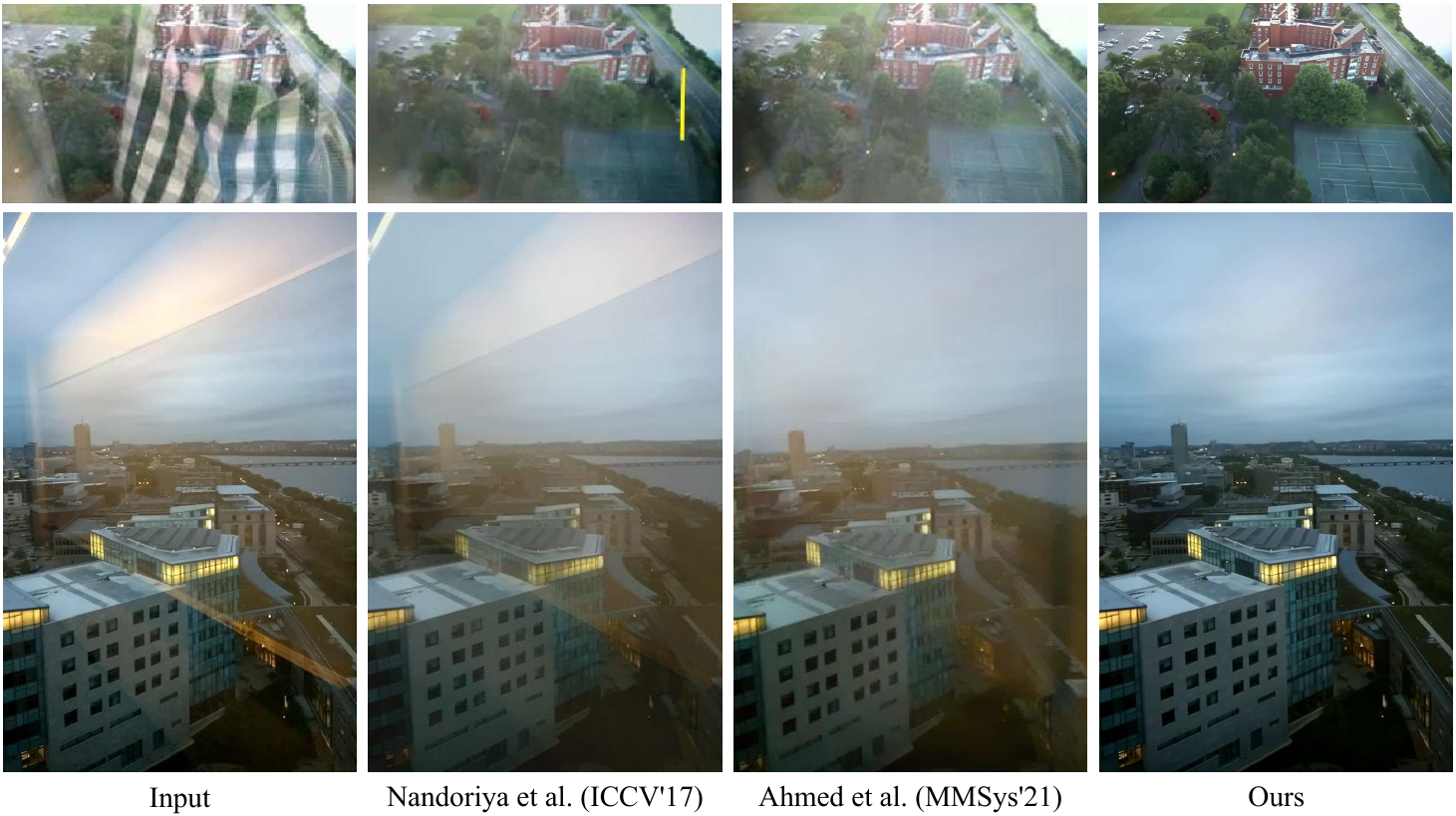}
    \caption{Qualitative comparison with prior video reflection removal methods~\cite{nandoriya2017video,ahmed2021user}. Their result frames are extracted from the original papers as no code is publicly available.}
    \label{fig:sup_frame_old_video}
\end{figure*}

\begin{figure*}[t]
    \centering
    \includegraphics[width=1.0\textwidth]{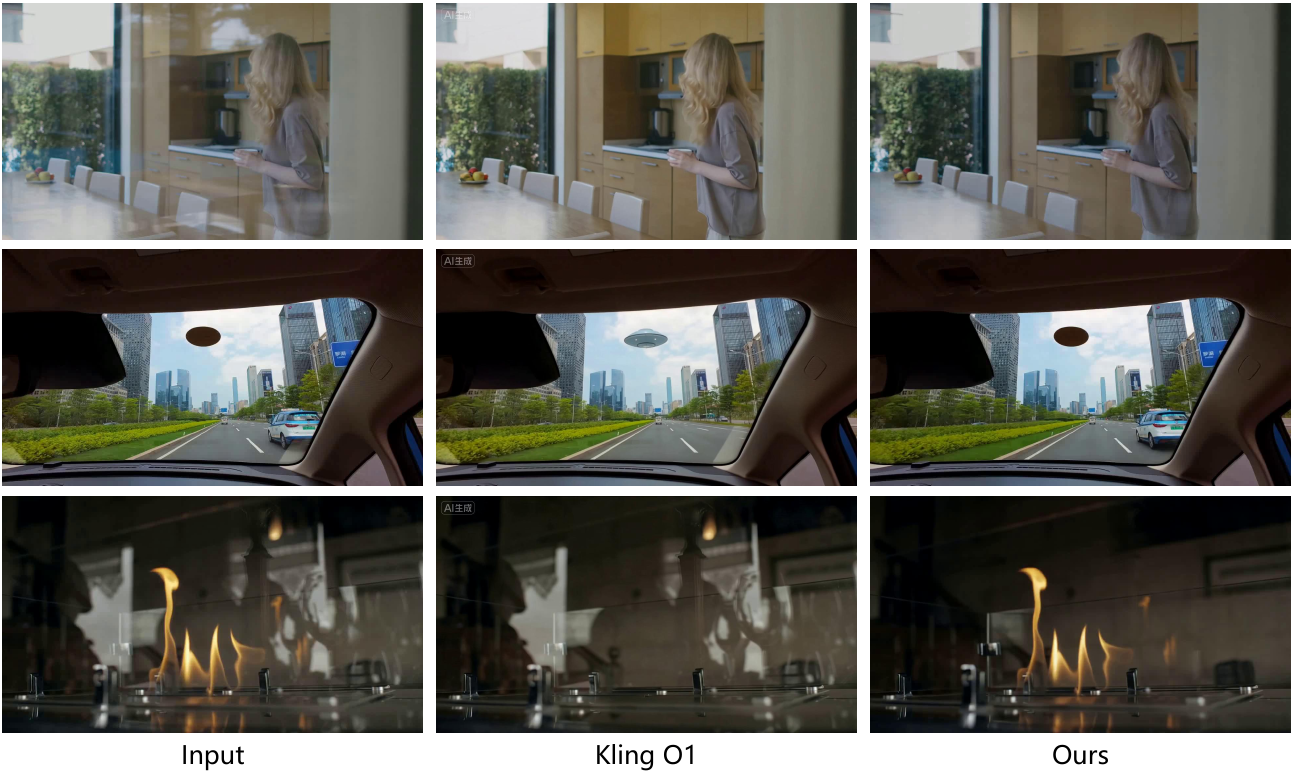}
    \caption{Qualitative comparison with the general-purpose video editing model Kling O1 on three representative cases.}
    \label{fig:sup_frame_kling}
\end{figure*}

\section{Additional Qualitative Comparisons}
\label{sec_sup:additional_qual}

We provide additional qualitative results complementing the main paper's SOTA comparison: a comparison on real-world reflection videos without ground truth (\Cref{fig:qual_real}), a qualitative comparison with prior video reflection removal methods (\Cref{fig:sup_frame_old_video}), and a comparison with a general-purpose video editing model (\Cref{fig:sup_frame_kling}).

\paragraph{Real-world qualitative comparison.}
\Cref{fig:qual_real} compares S2R-Removal with recent image reflection removal methods on real-world reflection videos without ground truth. Image-based methods often leave reflection residues or produce temporally inconsistent removal across frames, while S2R-Removal more completely suppresses strong reflections while preserving the original scene appearance.

\paragraph{Comparison with prior video dereflection methods.}
As noted in the main paper, prior video reflection removal methods~\cite{nandoriya2017video,ahmed2021user} release no code, precluding quantitative comparison; we therefore compare qualitatively on their respective test cases in~\Cref{fig:sup_frame_old_video}, with their result frames extracted directly from the original papers. S2R-Removal visibly outperforms these methods in both reflection suppression and transmission preservation, particularly on strong reflections where prior methods leave noticeable residues or alter the underlying scene content.

\paragraph{Comparison with general-purpose video editing models.}
We also compare against closed-source general-purpose video editing models. \Cref{fig:sup_frame_kling} compares S2R-Removal with Kling O1~\cite{team2025kling} on three representative cases. While general editing models show strong editing ability, they are not specifically trained for reflection layer separation and may alter the global appearance, scene content, or background structure. S2R-Removal performs more targeted dereflection and better preserves the original video content.

\section{Component Visualization}
\label{sec_sup:attmap}

\begin{figure*}[t]
    \centering
    \includegraphics[width=\textwidth]{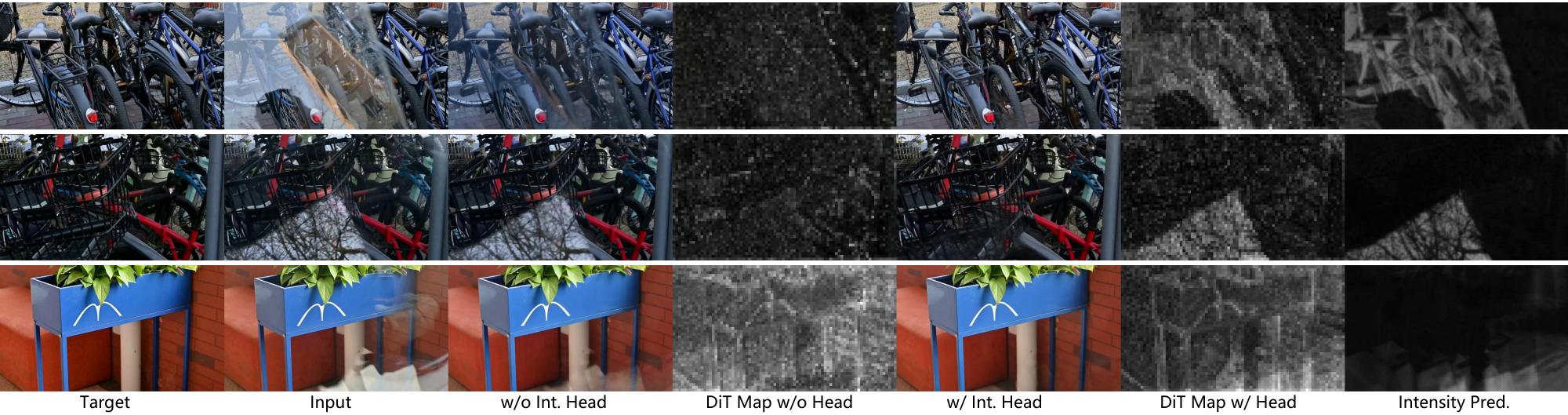}
    \caption{
    Visualization of the reflection-intensity head in Stage I.
    }
    \label{fig:intensity_head_vis}
\end{figure*}

\begin{figure*}[t]
    \centering
    \includegraphics[width=0.97\textwidth]{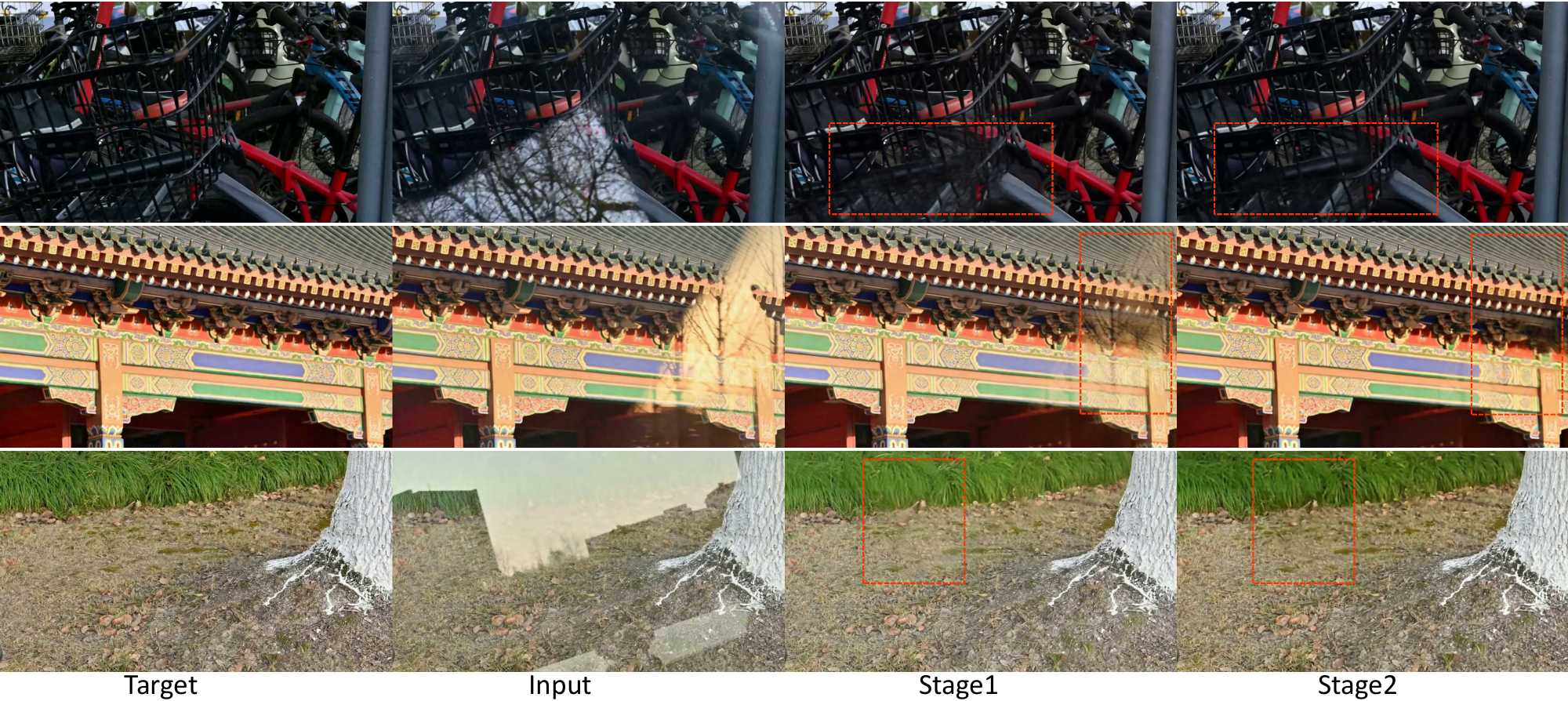}
    \caption{Qualitative comparison between Stage~I and Stage~II of S2R-Removal.}
    \label{fig:stage_refine_vis}
\end{figure*}

\subsection{Reflection-Intensity Head}
\label{sec_sup:attmap:intensity}

\paragraph{Attention Map Extraction.}
During inference, we extract the cross-attention weights from the DiT backbone of the Wan2.1 video diffusion model. Specifically, for each attention layer, the attention weights between visual Queries and textual Keys are computed as:
\begin{equation}
A = \text{softmax}\left(\frac{QK^\top}{\sqrt{d}}\right)
\in \mathbb{R}^{L_{vis} \times L_{text}},
\end{equation}
where $d$ denotes the attention head dimension, and $L_{vis}$ and $L_{text}$ denote the sequence lengths of visual tokens and text tokens, respectively.

We average the attention maps across all attention heads and Transformer layers to obtain the aggregated cross-attention map:
\begin{equation}
\bar{A}
=
\frac{1}{N_{layer}}
\sum_{l=1}^{N_{layer}}
\frac{1}{N_{head}}
\sum_{h=1}^{N_{head}}
A_l^h
\in
\mathbb{R}^{L_{vis} \times L_{text}}.
\end{equation}

\paragraph{Text Token Attention Response.}
We extract the attention response of all visual tokens to the text token ``reflection'' (indexed by $s$):
\begin{equation}
a_s = \bar{A}[:, s]
\in
\mathbb{R}^{L_{vis}}.
\end{equation}

This vector represents the response intensity of each spatio-temporal location in the video to the semantic concept ``reflection''. The vector is reshaped into a spatio-temporal grid $(F', H', W')$, where $F'$, $H'$, $W'$ denote the temporal and spatial dimensions in the latent space. It is then spatially upsampled and temporally interpolated to recover the full pixel-space resolution $(F, H, W)$, where:
\begin{equation}
F = 4(F' - 1) + 1,
\end{equation}
yielding the final frame-wise attention map:
\begin{equation}
\mathcal{A}
\in
\mathbb{R}^{F \times H \times W}.
\end{equation}
\paragraph{Visualization.}
The attention map is normalized into the range $[0, 255]$ for visualization:
\begin{equation}
\hat{A} = \frac{A - \min(A)}{\max(A)-\min(A)+\epsilon} \times 255.
\end{equation}

\Cref{fig:intensity_head_vis} visualizes the attention map activations with and without the intensity head. The columns show the clean target, reflected input, removal result without the intensity head, DiT response without the intensity head, removal result with the intensity head, DiT response with the intensity head, and the predicted reflection-intensity map. Without the intensity head, strong reflection regions are less attended by the DiT and remain difficult to remove; with residual-derived intensity supervision, the model produces more reflection-aware responses that better align with reflection-corrupted regions, and the predicted intensity map further captures both reflection location and strength, demonstrating the benefit of explicit reflection-intensity supervision.

\subsection{Stage~I vs Stage~II}
\label{sec_sup:attmap:stage}

\Cref{fig:stage_refine_vis} compares the two stages of S2R-Removal on representative reflection videos. Stage~I suppresses most reflection artifacts but may introduce appearance shifts or structural inconsistencies; Stage~II applies one-step pixel-geometric refinement to recover finer appearance details and better preserve scene geometry. The corresponding quantitative gains are reported in \Cref{tab:s2rdiff_ablation}.

\begin{figure*}[t]
    \centering
    \includegraphics[width=\textwidth]{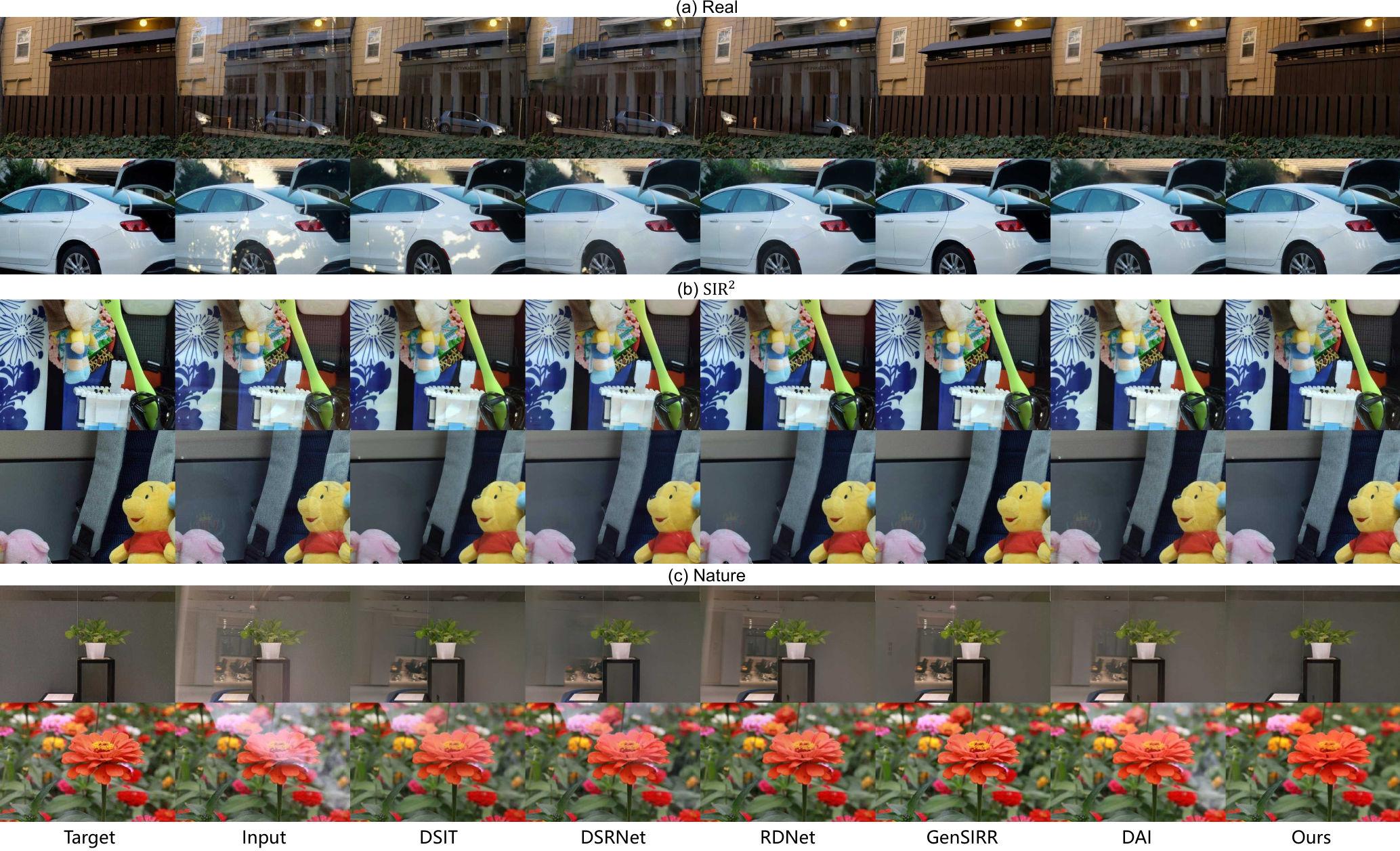}
    \caption{Qualitative comparison on the Real, SIR$^2$, and Nature image benchmarks.}
    \label{fig:imagebench_qual}
\end{figure*}

\section{Application to Single Images}
\label{sec_sup:image_eval}

\begin{table}[t]
\centering
\caption{Single-image inference vs. zoom-based video inference on image reflection removal benchmarks.}
\label{tab:image_eval}
\begin{tabular}{lcccccccc}
\toprule
\multirow{2}{*}{\textbf{Method}} & \multicolumn{2}{c}{\textbf{Real}} & \multicolumn{2}{c}{\textbf{SIR}$^2$} & \multicolumn{2}{c}{\textbf{Nature}} & \multicolumn{2}{c}{\textbf{Average}} \\
\cmidrule(lr){2-3} \cmidrule(lr){4-5} \cmidrule(lr){6-7} \cmidrule(lr){8-9}
& PSNR$\uparrow$ & SSIM$\uparrow$ & PSNR$\uparrow$ & SSIM$\uparrow$ & PSNR$\uparrow$ & SSIM$\uparrow$ & PSNR$\uparrow$ & SSIM$\uparrow$ \\
\midrule
Single-image & 27.61 & 0.857 & 28.27 & 0.888 & 27.45 & 0.841 & 28.21 & 0.885 \\
Zoom-in      & 27.85 & 0.866 & 28.62 & 0.899 & 27.58 & 0.857 & 28.55 & 0.896 \\
Zoom-out     & \textbf{28.23} & \textbf{0.881} & \textbf{28.83} & \textbf{0.925} & \textbf{27.89} & \textbf{0.883} & \textbf{28.77} & \textbf{0.921} \\
\bottomrule
\end{tabular}

\end{table}

Although our model is designed for video reflection removal, it can also be applied to single images by converting them into short video sequences.
Specifically, given a reflection-contaminated image, we synthesize a pseudo video clip by applying a smooth zoom transformation, generating $N$ frames via linear interpolation between scale factors.
We consider two zoom modes: zoom-in (scale from $1.0$ to $s$) and zoom-out (scale from $s$ to $1.0$), where $s=1.5$, $N=45$ frames.

Our model is then applied to the synthesized pseudo video clip; for zoom-in the first output frame is used for evaluation, and for zoom-out the last frame is used instead. \Cref{tab:image_eval} compares the three inference modes on the Real, SIR$^2$, and Nature benchmarks: zoom-based video inference consistently outperforms direct single-image inference, and zoom-out achieves the best results across all three benchmarks, which we therefore adopt for single-image evaluation.

Using zoom-out inference, we further compare S2R-Removal with state-of-the-art methods on the same image benchmarks (\Cref{fig:imagebench_qual}). Although trained for video dereflection, our model achieves more effective reflection removal while better preserving the transmission content, consistent with the quantitative cross-method results in~\Cref{tab:sota_combined}.

\section{Downstream Benefits of Dereflection}
\label{sec:downstream}

\begin{table}[t]
\centering
\caption{
Downstream driving area segmentation (YOLOP~\cite{wu2022yolop}) on reflection-contaminated BDD100K videos, before (Raw) and after (DeRef) our video reflection removal.
}
\label{tab:downstream_yolop}
\setlength{\tabcolsep}{4.0pt}
\renewcommand{\arraystretch}{1.1}
\begin{tabular}{llccc}
\toprule
{\textbf{Model}}
& {\textbf{Input}}
& \textbf{Acc$\uparrow$}
& \textbf{IoU$\uparrow$}
& \textbf{mIoU$\uparrow$}  \\
\midrule
\multirow{2}{*}{YOLOP}
& Raw
& 0.965 & 0.823 & 0.891\\
& DeRef
& \textbf{0.967} & \textbf{0.835} & \textbf{0.898}\\
\bottomrule
\end{tabular}

\end{table}

\begin{table}[t]
\centering
\caption{
Downstream vehicle detection (YOLOv5n~\cite{yolov5}) on reflection-contaminated BDD100K videos, before (Raw) and after (DeRef) our video reflection removal.
}
\label{tab:downstream_yolov5n}
\setlength{\tabcolsep}{4.0pt}
\renewcommand{\arraystretch}{1.1}
\begin{tabular}{llcccc}
\toprule
{\textbf{Model}}
& {\textbf{Input}}
& \textbf{P$\uparrow$}
& \textbf{R$\uparrow$}
& \textbf{mAP@0.5$\uparrow$}
& \textbf{mAP@0.95$\uparrow$} \\
\midrule
\multirow{2}{*}{YOLOv5n}
& Raw
& 0.827 & 0.361 & 0.351 & 0.221 \\
& DeRef
& \textbf{0.844} & \textbf{0.385} & \textbf{0.372} & \textbf{0.238} \\
\bottomrule
\end{tabular}

\end{table}

Reflection removal can serve as a useful preprocessing step for downstream vision systems by providing cleaner visual inputs.
To evaluate the practical benefit of our method, we conduct downstream task evaluation on reflection-contaminated videos from BDD100K~\cite{yu2020bdd100k}.
We compare task performance before and after applying our video dereflection model.
The evaluation covers two representative autonomous driving tasks: driving area segmentation and object detection.
We use YOLOP for joint driving perception, and YOLOv5n as an additional object detection model.

\begin{figure*}[t]
    \centering
    \begin{subfigure}[t]{0.48\linewidth}
        \centering
        \includegraphics[width=\linewidth]{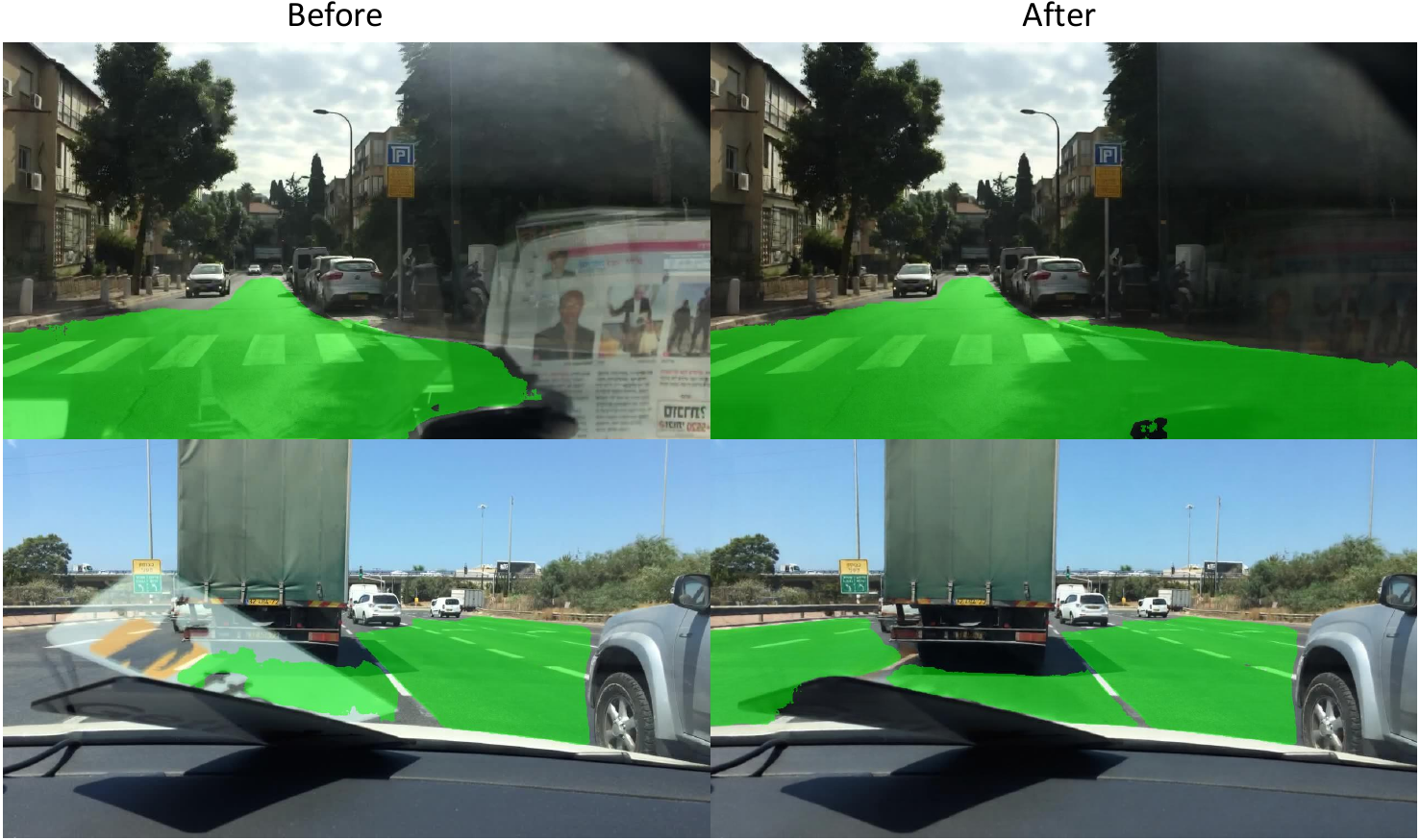}
        \caption{Driving area segmentation (YOLOP)}
        \label{fig:downstream_seg}
    \end{subfigure}
    \hfill
    \begin{subfigure}[t]{0.48\linewidth}
        \centering
        \includegraphics[width=\linewidth]{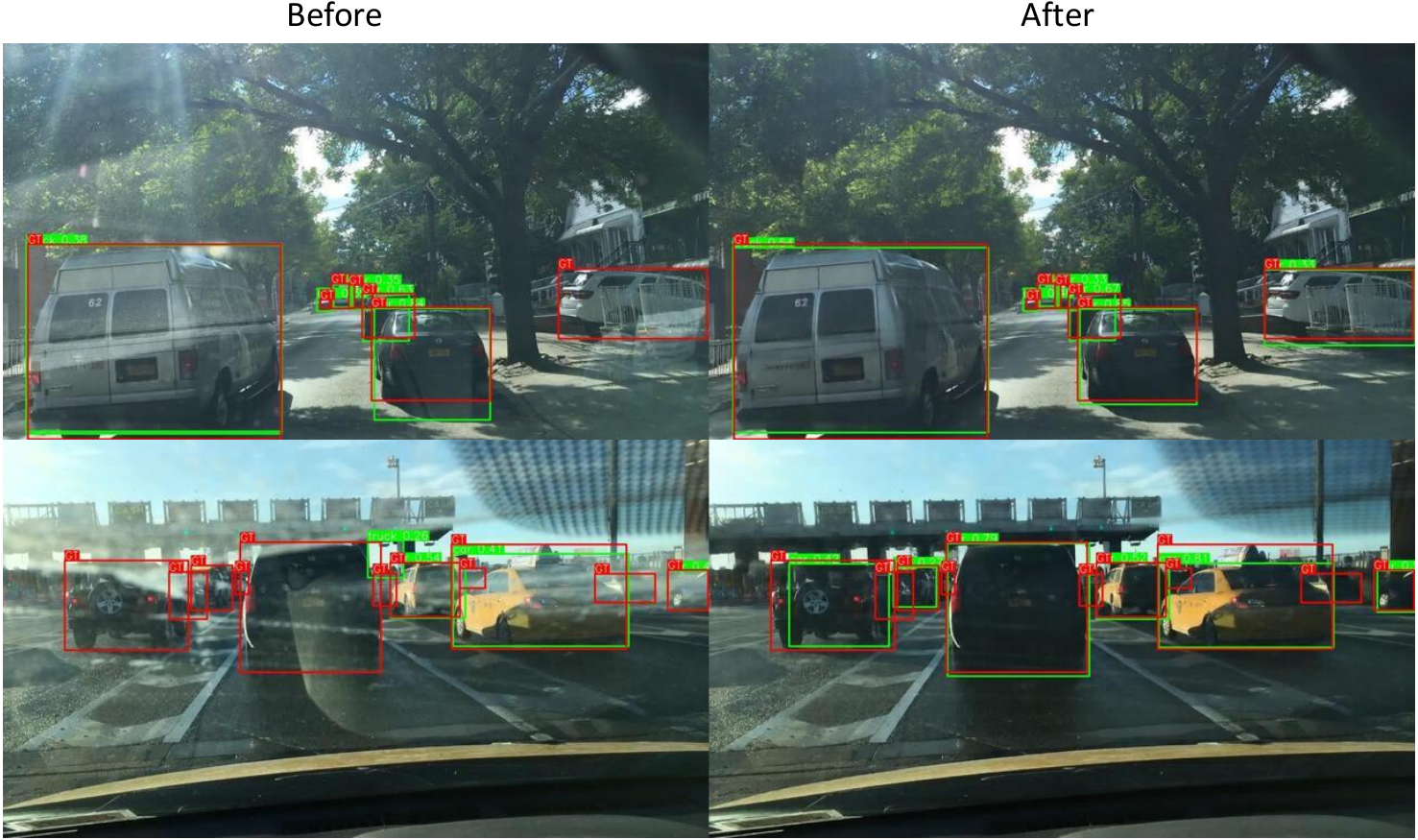}
        \caption{Object detection (YOLOv5n)}
        \label{fig:downstream_det}
    \end{subfigure}

    \caption{Qualitative results on downstream perception tasks.
    The proposed dereflection model improves visual clarity and benefits both segmentation and detection performance.}
    \label{fig:downstream_vis}
\end{figure*}

\Cref{tab:downstream_yolop,tab:downstream_yolov5n} report the quantitative results, while \Cref{fig:downstream_vis} presents representative qualitative examples.
After applying dereflection, YOLOP~\cite{wu2022yolop} improves driving area segmentation performance, with IoU increasing from 0.823 to 0.835 and mIoU increasing from 0.891 to 0.898.
As shown in \Cref{fig:downstream_seg}, removing reflections enables more complete and accurate segmentation of the drivable area.

For vehicle detection, YOLOv5n~\cite{yolov5} also achieves consistent improvements after dereflection, with mAP@0.5 increasing from 0.351 to 0.372 and mAP@0.95 increasing from 0.221 to 0.238.
The qualitative results in \Cref{fig:downstream_det} further demonstrate that dereflection effectively alleviates reflection-induced missed detections and improves vehicle localization accuracy.
Overall, our dereflection model effectively reduces reflection-induced interference and improves downstream perception in visually challenging scenarios.

\begin{figure*}[t]
    \centering
    \includegraphics[width=\textwidth]{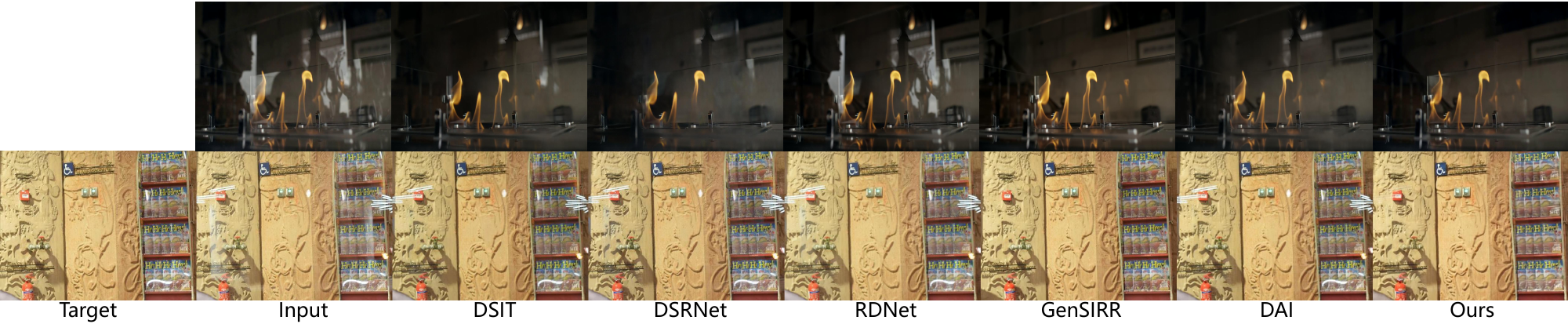}
    \caption{
Limitation in multi-layer reflection scenes. Due to ambiguous nested reflections, the model may leave slight inner-layer residuals (top, S2R-Real) or over-remove them (bottom, S2R-Ref).
}
    \label{fig:limitation_multilayer}
\end{figure*}

\section{Limitations}
\label{sec_sup:limitation}
Our current framework still has two limitations.

First, nested reflections from multiple glass layers remain challenging.
Since reflections behind another transparent surface may either be treated as removable artifacts or as part of the scene, the restoration target becomes ambiguous.
As illustrated in~\Cref{fig:limitation_multilayer}, our model may therefore either leave slight residual inner-layer reflections (top) or over-remove them (bottom).
This issue is also common to existing dereflection methods, and may be mitigated by incorporating more multi-layer reflection cases into future training data.

Second, our synthesis pipeline currently models temporally coherent reflections with clip-level controls, but does not explicitly simulate the coupled change between camera motion and reflection geometry, such as viewpoint-dependent reflection parallax.
Future work will explore richer physical reflection simulation and larger real-world video benchmarks to further improve robustness in complex glass scenarios.

\end{document}